\documentclass[11pt,letterpaper]{article}
\usepackage{emnlp2017}
\usepackage{times}
\usepackage{latexsym}

\emnlpfinalcopy

\def\emnlppaperid{1462}
\usepackage{floatrow} % Alex: has to go here in order to compile...

\usepackage{graphicx} % more modern
\usepackage{natbib}

\usepackage{algorithm}
\usepackage{algorithmic}

\usepackage{hyperref}

\usepackage{caption}
\usepackage{subcaption}
\usepackage{booktabs}
\usepackage{multirow}
\usepackage{amsmath}
\usepackage{amsfonts}
\usepackage{anyfontsize}
\usepackage{stfloats}

\begin{document} 

\title{Piecewise Latent Variables for Neural Variational Text Processing}

\author{Iulian V. Serban$^1$\thanks{ \hspace{0.1cm} The first two authors contributed equally.} \and Alexander G. Ororbia II$^2$\footnotemark[1] \and Joelle Pineau$^3$ \and Aaron Courville$^1$ \\ $^1$ Department of Computer Science and Operations Research, Universite de Montreal \\ $^2$College of Information Sciences \& Technology, Penn State University \\ $^3$School of Computer Science, McGill University \\ \texttt{iulian\hspace{0.1cm}[DOT]\hspace{0.1cm}vlad\hspace{0.1cm}[DOT]\hspace{0.1cm}serban\hspace{0.1cm}[AT]\hspace{0.1cm}umontreal\hspace{0.1cm}[DOT]\hspace{0.1cm}ca} \\ \texttt{ago109\hspace{0.1cm}[AT]\hspace{0.1cm}psu\hspace{0.1cm}[DOT]\hspace{0.1cm}edu} \\ \texttt{jpineau\hspace{0.1cm}[AT]\hspace{0.1cm}cs\hspace{0.1cm}[DOT]\hspace{0.1cm}mcgill\hspace{0.1cm}[DOT]\hspace{0.1cm}ca} \\ \texttt{aaron\hspace{0.1cm}[DOT]\hspace{0.1cm}courville\hspace{0.1cm}[AT]\hspace{0.1cm}umontreal\hspace{0.1cm}[DOT]\hspace{0.1cm}ca}}

\setlength\titlebox{2.50in}

%AO: I updated the title to reflect one reviewer's post-rebuttal comment
%    I also chose this title since I think it keeps things somewhat more general (since we train various encoder-decoder models) and we do start from general neural variational inference and work our way towards showing how the piecewise variables fit in and how the models change as a result :)
\maketitle

\begin{abstract}
Advances in neural variational inference have facilitated the learning of powerful directed graphical models with continuous latent variables, such as variational autoencoders.
%The hope is that such models will learn to represent latent factors in a wide range of real-world data.
The hope is that such models will learn to represent rich, multi-modal latent factors in real-world data, such as natural language text.
%However, latent factors in real-world data --- such as natural language text -- often posses latent factors, which follow highly non-linear, multi-modal distributions.
%, such as natural language text. %, audio, and images.
%However, latent factors in real-world data are often highly complex; for example, topics in newswire text and responses in conversational dialogue often posses latent factors, which follow non-linear (non-smooth), multi-modal distributions. % with many statistical outliers.
However, current models often assume simplistic priors on the latent variables --- such as the uni-modal Gaussian distribution --- which are incapable of representing complex latent factors efficiently. To overcome this restriction, we propose the simple, but highly flexible, piecewise constant distribution.
This distribution has the capacity to represent an exponential number of modes of a latent target distribution, while remaining mathematically tractable. Our results demonstrate that incorporating this new latent distribution into different models yields substantial improvements in natural language processing tasks such as document modeling and natural language generation for dialogue.
\end{abstract}

% OLD ABSTRACT
% \begin{abstract}
% Recent advances in neural variational inference have facilitated training of powerful directed graphical models with continuous latent variables, such as variational autoencoders.
% However, these models usually assume simple priors --- such as the uni-modal Gaussian distribution --- yet many real-world data distributions are highly complex and multi-modal.
% Examples of complex, multi-modal distributions range from topics in newswire text to conversational dialogue responses.
% When such latent variable models are applied to these domains, the restrictive priors make it difficult to capture non-linear, multi-modal probability manifolds over latent concepts.
% To overcome this critical restriction, we propose the simple, but highly flexible, piecewise constant distribution.
% The piecewise constant distribution has the capacity to represent an exponential number of modes of a latent target distribution, and --- as we will show --- it is mathematically tractable within the variational autoencoder framework.
% %In comparison to alternative approaches, such as stacked multivariate Gaussian distributions or auto-regressive approximate posteriors, our latent distribution does not require additional approximations or multiple sampling steps.
% We show that incorporating this new latent variable parametrization into neural models yields substantial performance improvements in natural language processing tasks such as document modeling and dialogue modeling.
% \end{abstract} 

\section{Introduction}
\label{intro}
The development of the variational autoencoder framework \citep{kingma2013auto,rezende2014stochastic} has paved the way for learning large-scale, directed latent variable models. This has led to significant progress in a diverse set of machine learning applications, ranging from computer vision \citep{gregor2015draw,larsen2015autoencoding} to natural language processing tasks \citep{mnih2014neural,miao2015neural,bowman2015generating,serban2016hierarchical}.
It is hoped that this framework will enable the learning of generative processes of real-world data --- including text, audio and images --- by disentangling and representing the underlying latent factors in the data. However, latent factors in real-world data are often highly complex. For example, topics in newswire text and responses in conversational dialogue often posses latent factors that follow non-linear (non-smooth), multi-modal distributions (i.e.\@ distributions with multiple local maxima).

Nevertheless, the majority of current models assume a simple prior in the form of a multivariate Gaussian distribution in order to maintain mathematical and computational tractability.
This is often a highly restrictive and unrealistic assumption to impose on the structure of the latent variables.
First, it imposes a strong uni-modal structure on the latent variable space; latent variable samples from the generating model (prior distribution) all cluster around a single mean.
Second, it forces the latent variables to follow a perfectly symmetric distribution with constant kurtosis; this makes it difficult to represent asymmetric or rarely occurring factors.
%Second, it encourages local smoothness on the latent variables, where the similarity between two latent variable samples decreases exponentially as their distance increase.
%Furthermore, variational autoencoding models naturally incorporate a Bayesian modeling perspective,
%by enabling the integration of problem-dependent knowledge within the prior on the generating distribution.
Such constraints on the latent variables increase pressure on the down-stream generative model, which in turn is forced to carefully partition the probability mass for each latent factor throughout its intermediate layers.
For complex, multi-modal distributions --- such as the distribution over topics in a text corpus, or natural language responses in a dialogue system --- the uni-modal Gaussian prior inhibits the model's ability to extract and represent important latent structure in the data.
In order to learn more expressive latent variable models, we therefore need more flexible, yet tractable, priors.

In this paper, we introduce a simple, flexible prior distribution based on the piecewise constant distribution.
We derive an analytical, tractable form that is applicable to the variational autoencoder framework and propose a differentiable parametrization for it.
We then evaluate the effectiveness of the distribution when utilized both as a prior and as approximate posterior across variational architectures in two natural language processing tasks: document modeling and natural language generation for dialogue.
We show that the piecewise constant distribution is able to capture elements of a target distribution that cannot be captured by simpler priors --- such as the uni-modal Gaussian.
We demonstrate state-of-the-art results on three document modeling tasks, and show improvements on a dialogue natural language generation.
Finally, we illustrate qualitatively how the piecewise constant distribution represents multi-modal latent structure in the data.

\section{Related Work}
\label{related_work}
The idea of using an artificial neural network to approximate an inference model dates back to the early work of Hinton and colleagues \citep{hinton_mdl_1994,hinton1995wake,dayan1996varieties}.
%However, initial work lacked low-bias, low-variance estimators for the parameter updates. 
Researchers later proposed Markov chain Monte Carlo methods (MCMC) \citep{neal1992connectionist}, which do not scale well and mix slowly, as well as variational approaches which require a tractable, factored distribution to approximate the true posterior distribution \citep{jordan1999introduction}.
Others have since proposed using feed-forward inference models to initialize the mean-field inference algorithm for training Boltzmann architectures \citep{salakhutdinov2010efficient,ororbia2015online}.
%However, these approaches are limited by the mean-field inference's inability to model structured posteriors.
Recently, the variational autoencoder framework (VAE) was proposed by \citet{kingma2013auto} and \citet{rezende2014stochastic}, closely related to the method proposed by \citet{mnih2014neural}.
This framework allows the joint training of an inference network and a directed generative model, maximizing a variational lower-bound on the data log-likelihood and facilitating exact sampling of the variational posterior. 
Our work extends this framework. %in the next section.

With respect to document modeling, neural architectures have been shown to outperform well-established topic models such as Latent Dirichlet Allocation (LDA) \citep{hofmann1999probabilistic,blei2003latent}.
%For example, it has been demonstrated that models based on the Boltzmann machine, which learn semantic binary vectors (binary latent variables), perform very well\citep{hofmann1999probabilistic}.
Researchers have successfully proposed several models involving discrete latent variables \citep{salakhutdinov2009semantic,hinton_softmax_2009,srivastava2013modeling,larochelle_neural_2012,uria2014deep,lauly2016document,bornschein2014reweighted,mnih2014neural}.
%Work involving discrete latent variables include the constrained Poisson model \citep{salakhutdinov2009semantic}, the Replicated Softmax model \citep{hinton_softmax_2009} and the Over-Replicated Softmax model \citep{srivastava2013modeling}, as well as similar, auto-regressive neural architectures and deep directed graphical models \citep{larochelle_neural_2012,uria2014deep,lauly2016document,bornschein2014reweighted}.
%In particular, \citet{mnih2014neural} showed that using NVIL yields excellent generative models of documents.
The success of such discrete latent variable models --- which are able to partition probability mass into separate regions --- serves as one of our main motivations for investigating models with more flexible continuous latent variables for document modeling.
% , multi-modal, continuous latent variables for document modeling.
More recently, \citet{miao2015neural} proposed to use continuous latent variables for document modeling. %achieving state-of-the-art results.
%This model will be described later.

Researchers have also investigated latent variable models for dialogue modeling and dialogue natural language generation \citep{bangalore2008learning, crook2009unsupervised,zhai2014discovering}.
The success of discrete latent variable models in this task also motivates our investigation of more flexible continuous latent variables.
Closely related to our proposed approach is the Variational Hierarchical Recurrent Encoder-Decoder (\emph{VHRED}, described below) \citep{serban2016hierarchical}, a neural architecture with latent multivariate Gaussian variables.
In parallel with our work, \citet{zhao2017learning} has also proposed a latent variable model for dialogue modeling with the specific goal of generating diverse natural language responses.

Researchers have explored more flexible distributions for the latent variables in VAEs, such as autoregressive distributions, hierarchical probabilistic models and approximations based on MCMC sampling \citep{rezende2014stochastic,rezende2015variational,kingma2016improving,ranganath2016hierarchical,maaloe2016auxiliary,salimans2015markov,burda2015importance,chen2016variational,ruiz2016generalized}.
These are all complimentary to our approach; it is possible to combine them with the piecewise constant latent variables.
In parallel to our work, multiple research groups have also proposed VAEs with discrete latent variables \citep{maddison2016concrete,jang2016categorical,rolfe2016discrete,johnson2016composing}.
This is a promising line of research, however these approaches often require approximations which may be inaccurate when applied to larger scale tasks, such as document modeling or natural language generation.
Finally, discrete latent variables may be inappropriate for certain natural language processing tasks.

\section{Neural Variational Models}
We start by introducing the neural variational learning framework.
%Then, we present the proposed piecewise constant distribution, which aims to be a tractable, highly flexible distribution for the model prior and approximate posterior.
We focus on modeling discrete output variables (e.g.\@ words) in the context of natural language processing applications.
However, the framework can easily be adapted to handle continuous output variables.%, such as images.

\subsection{Neural Variational Learning}
Let $w_1, \dots, w_N$ be a sequence of $N$ tokens (words) conditioned on a continuous latent variable $z$.
Further, let $c$ be an additional observed variable which conditions both $z$ and $w_1, \dots, w_N$.
Then, the distribution over words is:
{\fontsize{10}{12} % Alex: I used fontsize re-scaling, not sure if there's a better way :-(
\begin{align}
P_{\theta}(w_1, \dots, w_N | c) = \int \prod_{n=1}^N P_{\theta}(w_n | w_{<n}, z, c) P_{\theta}(z | c) dz, \nonumber
\end{align}
}%
\noindent
where $\theta$ are the model parameters.
The model first generates the higher-level, continuous latent variable $z$ conditioned on $c$.
Given $z$ and $c$, it then generates the word sequence $w_1, \dots, w_N$.
For unsupervised modeling of documents, the $c$ is excluded and the words are assumed to be independent of each other, when conditioned on $z$:
{\fontsize{10}{12}
\begin{align}
P_{\theta}(w_1, \dots, w_N) = \int \prod_{n=1}^N P_{\theta}(w_n | z) P_{\theta}(z) dz. \nonumber
\end{align}
}
Model parameters can be learned using the variational lower-bound \citep{kingma2013auto}:
{\fontsize{10}{12}
\begin{align}
\label{eq:variatonal_lower_bound}
\log P_{\theta} & (w_1, \dots, w_N | c) \nonumber \\
 \geq & \quad \text{E}_{z \sim Q_{\psi}(z | w_1, \dots, w_N, c) }[\log P_{\theta}(w_n | w_{<n}, z, c)] \nonumber \\
     & - \text{KL} \left [ Q_{\psi}(z | w_1, \dots, w_N, c) || P_{\theta}(z | c) \right ], 
\end{align}
}
%\begin{align}
%\log P_{\theta}(w_1, \dots, w_N, z)  \geq \text{E}_{z \sim Q_{\psi}(z | w_1, \dots, w_N) }[\log P_{\theta}(w_n | w_{<n}, z)] - \text{KL} \left [ Q_{\psi}(z | w_1, \dots, w_N) || P_{\theta}(z) \right ], \label{eq:variatonal_lower_bound}
%\end{align}
\noindent 
where we note that $Q_{\psi}(z | w_1, \dots, w_N, c)$ is the approximation to the intractable, true posterior $P_{\theta}(z | w_1, \dots, w_N, c)$.
$Q$ is called the \textit{encoder}, or sometimes the \textit{recognition model} or \textit{inference model}, and it is parametrized by $\psi$.
The distribution $P_{\theta}(z | c)$ is the prior model for $z$, where the only available information is $c$.
The VAE framework further employs the re-parametrization trick, which allows one to move the derivative of the lower-bound inside the expectation.
To accomplish this, $z$ is parametrized as a transformation of a fixed, parameter-free random distribution $z = f_\theta(\epsilon)$,
%\begin{align}
%z = f_\theta(\epsilon),
%\end{align}
where $\epsilon$ is drawn from a random distribution. %(e.g.\@ a standard Gaussian distribution with zero mean and unit standard deviation, or a uniform distribution in the interval $[0, 1]$).
Here, $f$ is a transformation of $\epsilon$, parametrized by $\theta$, such that $f_\theta(\epsilon) \sim P_{\theta}(z | c)$.
For example, $\epsilon$ might be drawn from a standard Gaussian distribution and $f$ might be defined as $f_\theta(\epsilon) = \mu + \sigma \epsilon$, where $\mu$ and $\sigma$ are in the parameter set $\theta$.
In this case, $z$ is able to represent any Gaussian with mean $\mu$ and variance $\sigma^2$.

Model parameters are learned by maximizing the variational lower-bound in eq.\@ \eqref{eq:variatonal_lower_bound} using gradient descent, where the expectation is computed using samples from the approximate posterior. %See also \citet {kingma2013auto}.

The majority of work on VAEs propose to parametrize $z$ as multivariate Gaussian distribtions.
However, this unrealistic assumption may critically hurt the expressiveness of the latent variable model.
See Appendix \ref{appendix_0} for a detailed discussion. This motivates the proposed piecewise constant latent variable distribution.

% Aaron: \paragraph{Gaussian priors are sometimes inappropriate.} Here being concrete is important. You do so with your Ubuntu example, but you need to be more explicit about how a Gaussian would get it wrong. The issue you raise is an important one: that in a dialogue, conditioning on context could easily result in a complicated prior as there may be genuine ambiguities that are not easily resolvable with a unimodal variable. 
%\subsubsection{Gaussian priors are sometimes inappropriate}

\subsection{Piecewise Constant Distribution}
\label{piecewise_prior}
We propose to learn latent variables by parametrizing $z$ using a piecewise constant probability density function (PDF).
This should allow $z$ to represent complex aspects of the data distribution in latent variable space, such as non-smooth regions of probability mass and multiple modes.
%In particular, it should enable $z$ to represent islands of probability mass relevant both for representing diverse topics in news articles and for representing ambiguity and uncertainty in natural language conversations.
%From a manifold learning perspective, this extension translates into expanding the set of manifolds representable by the model parameters to include more non-linear manifolds -- in particular, manifolds where there exists separate clusters of probability mass.

\begin{figure}[!t]
\centering
\includegraphics[scale=0.35]{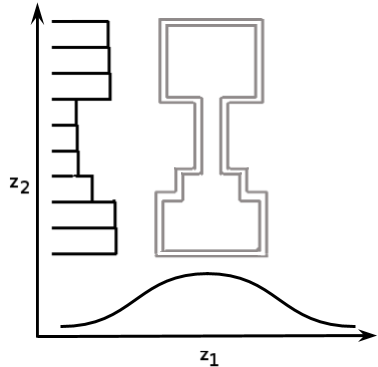}
\caption{Joint density plot of a pair of Gaussian and piecewise constant variables. The horizontal axis corresponds to $z_1$, which is a univariate Gaussian variable. The vertical axis corresponds to $z_2$, which is a piecewise constant variable.} %The density for each variable is shown along each axis, and their joint distribution is illustrated in grey color.}
% * <aaron.courville@gmail.com> 2016-11-10T18:33:52.821Z:
%
% I change the caption from x_1, x_2 to z_1, z_2 to match the axis labels in the fig.
%
% ^ <aaron.courville@gmail.com> 2016-11-16T16:17:05.244Z.
\label{fig:joint_prob}
\end{figure}

Let $n \in \mathbb{N}$ be the number of piecewise constant components. We assume $z$ is drawn from PDF:
{\fontsize{9.75}{11.7}
\begin{align}
P(z) = \dfrac{1}{K} \sum_{i=1}^n 1_{ \left (\dfrac{i-1}{n} \leq z \leq \dfrac{i}{n} \right )} a_i,\label{piecewise_pdf}
\end{align}
}
where $1_{(x)}$ is the indicator function, which is one when $x$ is true and otherwise zero.
The distribution parameters are $a_i > 0$, for $i=1,\dots,n$. The normalization constant is:
{\fontsize{9.75}{11.7}
\begin{align}
K & = \sum_{i=1}^n K_i, \ \text{where} \ K_0=0, K_i = \dfrac{a_i}{n}, \ \text{for} \ i=1,\dots,n. \nonumber
\end{align}
}
It is straightforward to show that a piecewise constant distribution with more than $n>2$ pieces is capable of representing a bi-modal distribution.
When $n>2$, a vector $z$ of piecewise constant variables can represent a probability density with $2^{|z|}$ modes.
Figure \ref{fig:joint_prob} illustrates how these variables help model complex, multi-modal distributions. %work with Gaussian latent variables in order to 

In order to compute the variational bound, we need to draw samples from the piecewise constant distribution using its inverse cumulative distribution  function (CDF).
Further, we need to compute the KL divergence between the prior and posterior.
The inverse CDF and KL divergence quantities are both derived in Appendix \ref{appendix_a}.
During training we must compute derivatives of the variational bound in eq.\@ \eqref{eq:variatonal_lower_bound}.
These expressions involve derivatives of indicator functions, which have derivatives zero everywhere except for the changing points where the derivative is undefined.
However, the probability of sampling the value exactly at its changing point is effectively zero. Thus, we fix these derivatives to zero. Similar approximations are used in training networks with rectified linear units. %\footnote{We thank Christian A.\@ Naesseth for pointing this out.}

\section{Latent Variable Parametrizations}
\label{latent_params}
%The parametrization of the latent variables is crucial for learning.
In this section, we develop the parametrization of both the Gaussian variable and our proposed piecewise constant latent variable. %, which is critical for efficient learning.

Let $x$ be the current output sequence, which the model must generate (e.g.\@ $w_1, \dots, w_N$).
Let $c$ be the observed conditioning information.
%for the prior.
If the task contains additional conditioning information this will be embedded by $c$. For example, for dialogue natural language generation $c$ represents an embedding of the dialogue history, while for document modeling $c=\emptyset$.
% If no conditioning information is available, then  
%In document modeling there is no conditioning information available to the prior, so $c=\emptyset$.
%In dialogue modeling $c$ is the vector representation of the dialogue context, namely all previous utterances until the current time step.

\subsection{Gaussian Parametrization}
\label{gaussian_param}

Let $\mu^{\text{prior}}$ and $\sigma^{2,\text{prior}}$ be the prior mean and variance, and let $\mu^{\text{post}}$ and $\sigma^{2,\text{post}}$ be the approximate posterior mean and variance.
For Gaussian latent variables, the prior distribution mean and variances are encoded using linear transformations of a hidden state.
In particular, the prior distribution covariance is encoded as a diagonal covariance matrix using a softplus function:
% TODO: Consider dropping equation numbers here! This will save space, and help reduce the cognitive load on reviewers..
{\fontsize{9.75}{11.7}
\begin{align}
\mu^{\text{prior}} &= H_{\mu}^{\text{prior}} \text{Enc}(c) + b_{\mu}^{\text{prior}}, \nonumber \\ % \label{eq:mu_prior} \\
\sigma^{2,\text{prior}} &= \text{diag}(\log(1 + \exp ( H_{\sigma}^{\text{prior}} \text{Enc}(c) + b_{\sigma}^{\text{prior}}))), \nonumber % \label{eq:sigma_prior}
\end{align}
}%
where $\text{Enc}(c)$ is an embedding of the conditioning information $c$ (e.g.\@ for dialogue natural language generation this might, for example, be produced by an LSTM encoder applied to the dialogue history), which is shared across all latent variable dimensions. 
The matrices $H_{\mu}^{\text{prior}}, H_{\sigma}^{\text{prior}}$ and vectors $b_{\mu}^{\text{prior}}, b_{\sigma}^{\text{prior}}$ are learnable parameters.
For the posterior distribution, previous work has shown it is better to parametrize the posterior distribution as a linear interpolation of the prior distribution mean and variance and a new estimate of the mean and variance based on the observation $x$ \citep{fraccaro2016sequential}.
The interpolation is controlled by a gating mechanism, allowing the model to turn on/off latent dimensions:
{\fontsize{9.75}{11.7}
\begin{align}
\mu^{\text{post}} = & (1 - \alpha_{\mu}) \mu^{\text{prior}} + \alpha_{\mu} \left ( H_{\mu}^{\text{post}} \text{Enc}(c, x) + b_{\mu}^{\text{post}} \right ), \nonumber \\ %\label{eq:mu_post}\\
\sigma^{2,\text{post}} = & (1 - \alpha_{\sigma}) \sigma^{2,\text{prior}} \nonumber \\ & + \alpha_{\sigma} \text{diag}(\log(1 + \exp ( H_{\sigma}^{\text{post}} \text{Enc}(c, x) + b_{\sigma}^{\text{post}}))), \nonumber % \label{eq:sigma_post}
\end{align}
}%
where $\text{Enc}(c, x)$ is an embedding of both $c$ and $x$.
The matrices $H_{\mu}^{\text{post}}, H_{\sigma}^{\text{post}}$ and the vectors $b_{\mu}^{\text{post}}, b_{\sigma}^{\text{post}}, \alpha_{\mu}, \alpha_{\sigma}$ are parameters to be learned.
The interpolation mechanism is controlled by $\alpha_{\mu}$ and $\alpha_{\sigma}$, which are initialized to zero (i.e.\@ initialized such that the posterior is equal to the prior). %However, we found that the simpler was often better and thus do not report these results using more advanced mechanisms.}
%Optionally, $\alpha_{\mu}$ and $\alpha_{\sigma}$ can be defined as a linear function of $\text{Enc}(c, x)$. This should be even better

\subsection{Piecewise Constant Parametrization}
\label{piecewise_param}

We parametrize the piecewise prior parameters using an exponential function applied to a linear transformation of the conditioning information:
\begin{align}
a_i^{\text{prior}} = \exp ( H_{a,i}^{\text{prior}} \text{Enc}(c) + b_{a,i}^{\text{prior}}), \quad i=1,\dots,n, \nonumber% \label{eq:a_i_prior}
\end{align}
%\begin{align}
%a_i^{\text{prior}} = \log(1 + \exp ( H_{a,i}^{\text{prior}} \text{Enc}(c) + b_{a,i}^{\text{prior}})), \quad i=1,\dots,n, \label{eq:a_i_prior}
%\end{align}
where matrix $H_{a}^{\text{prior}}$ and vector $b_{a}^{\text{prior}}$ are learnable.
As before, we define the posterior parameters as a function of both $c$ and $x$:
{\fontsize{9.75}{11.7}
\begin{align}
a_i^{\text{post}} = & \exp ( H_{a,i}^{\text{post}} \text{Enc}(c,x) + b_{a,i}^{\text{post}}), \quad i=1,\dots,n, \nonumber% \label{eq:a_i_posterior}
\end{align}
}
where $H_{a}^{\text{post}}$ and $b_{a}^{\text{post}}$ are parameters.
%However, we found that this interpolation hurt performance and therefore fixed $\alpha_{a} = \mathbf{1}$.

%We may also constrain the piecewise constant posterior parameters to be an interpolation between the prior parameters and a new estimated parameter:
%\begin{align}
%a_i^{\text{post}} = & (1 - \alpha_{a,i}) a_i^{\text{prior}} \nonumber \\ & + \alpha_{a,i} \exp ( H_{a,i}^{\text{post}} \text{Enc}(c,x) + b_{a,i}^{\text{post}}), \nonumber \\ & \quad i=1,\dots,n, \nonumber% \label{eq:a_i_posterior}
%\end{align}
%where $H_{a}^{\text{post}}, b_{a}^{\text{post}}, \alpha_{a}$ are the parameters.
%However, we found that this interpolation hurt performance and therefore fixed $\alpha_{a} = \mathbf{1}$.
%with an initialization of $\alpha_{a} = \mathbf{0}$.

%To take advantage of the representational powers of each prior, we may concatenate the Gaussian and piecewise constant variables into a single vector of latent variables.

%of each priors, the Gaussian and piecewise constant variables may be combined, as was suggested in Section \ref{piecewise_prior}. In this work, we primarily experimented with their concatenation to create a hybrid model.

\section{Variational Text Modeling}
We now introduce two classes of VAEs. The models are extended by incorporating the Gaussian and piecewise latent variable parametrizations.%, and are used for our experiments on document modeling and the dialogue modeling.

%\subsection{Neural Variational Document Model (NVDM)}
\subsection{Document Model}
\label{nvdm}
The neural variational document model  (\emph{NVDM}) model has previously been proposed for document modeling \citep{mnih2014neural,miao2015neural}, where the latent variables are Gaussian.
Since the original \emph{NVDM} uses Gaussian latent variables, we will refer to it as \emph{G-NVDM}.
We propose two novel models building on \emph{G-NVDM}.
The first model we propose uses piecewise constant latent variables instead of Gaussian latent variables.
We refer to this model as \emph{P-NVDM}.
The second model we propose uses a combination of Gaussian and piecewise constant latent variables.
The models sample the Gaussian and piecewise constant latent variables independently and then concatenates them together into one vector.
We refer to this model as \emph{H-NVDM}.
% Since this is a hybrid model, 

Let $V$ be the vocabulary of document words.
Let $W$ represent a document matrix, where row $w_i$ is the 1-of-$|V|$ binary encoding of the $i$'th word in the document.
Each model has an encoder component $Enc(W)$,
which compresses a document vector into a continuous distributed representation upon which the approximate posterior is built.
For document modeling, word order information is not taken into account and no additional conditioning information is available.
Therefore, each model uses a bag-of-words encoder, defined as a multi-layer perceptron (MLP) $Enc(c = \emptyset, x) = Enc(x)$.
Based on preliminary experiments, we choose the encoder to be a two-layered MLP with parametrized rectified linear activation functions (we omit these parameters for simplicity). % defined by parameters  $\{E^0,b^0,E^1,b^1\}$.
For the approximate posterior, each model has the parameter matrix $W_{a}^{\text{post}}$ and vector $b_{a}^{\text{post}}$ for the piecewise latent variables, and the parameter matrices $W_{\mu}^{\text{post}}, W_{\sigma}^{\text{post}}$ and vectors $b_{\mu}^{\text{post}}, b_{\sigma}^{\text{post}}$ for the Gaussian means and variances.
For the prior, each model has parameter vector $b_{a}^{\text{prior}}$ for the piecewise latent variables, and vectors $b_{\mu}^{\text{prior}}, b_{\sigma}^{\text{prior}}$ for the Gaussian means and variances.
%and approximate posterior distributions, each model requires learning the parameters, , b_{a}^{\text{post}}\}$ for the piecewise variables, and learning the parameters  for the Gaussian variables.
% Julian: I changed this, because only the posterior depends on $x$ for the NVDM model.
%$Enc(x)$ is trained to compress a document vector into a continuous distributed representation upon which the prior and posterior models are built.
%The NVDM parametrization simplifies the prior and posterior distribution models described for both the Gaussian and piecewise constant latent variables, requiring that only the bias parameters, $b_{a}^{\text{prior}}, b_{a}^{\text{post}}$ for the piecewise and  $b_{\mu}^{\text{post}}, b_{\sigma}^{\text{post}}, b_{\mu}^{\text{prior}}, b_{\sigma}^{\text{prior}}$ for the Gaussian, are learned.
We initialize the bias parameters to zero in order to start with centered Gaussian and piecewise constant priors.
The encoder will adapt these priors as learning progresses, using the gating mechanism to turn on/off latent dimensions.

Let $z$ be the vector of latent variables sampled according to the approximate posterior distribution.
Given $z$, the decoder $Dec(w, z)$ outputs a distribution over words in the document:
%\begin{align*}
%Dec(w, z) = P_{\theta}(w|z) = \frac{\exp{( -w^{\text{T}} R z + b_w)}}{\sum_{w'} \exp{(-w^{\text{T}} R z + b_{w'})}},
%\end{align*}
\begin{align*}
Dec(w, z) = \frac{\exp{( -w^{\text{T}} R z + b_w)}}{\sum_{w'} \exp{(-w^{\text{T}} R z + b_{w'})}},
\end{align*}
where $R$ is a parameter matrix and $b$ is a parameter vector corresponding to the bias for each word to be learned.
This output probability distribution is combined with the KL divergences to compute the lower-bound in eq.\@ \eqref{eq:variatonal_lower_bound}.
See Appendix \ref{appendix_c}. % for further details.

Our baseline model \emph{G-NVDM} is an improvement over the original \emph{NVDM} proposed by \citet{mnih2014neural} and \citet{miao2015neural}.
We learn the prior mean and variance, while these were fixed to a standard Gaussian in previous work. This increases the flexibility of the model and makes optimization easier.
In addition, we use a gating mechanism for the approximate posterior of the Gaussian variables.
This gating mechanism allows the model to turn off latent variable (i.e.\@ fix the approximate posterior to equal the prior for specific latent variables) when computing the final posterior parameters.
Furthermore, \citet{miao2015neural} alternated between optimizing the approximate posterior parameters and the generative model parameters, while we optimize all parameters simultaneously.
\subsection{Dialogue Model}
\label{vhred}

The variational hierarchical recurrent encoder-decoder (\emph{VHRED}) model has previously been proposed for dialogue modeling and natural language generation~\citep{serban2016hierarchical,DBLP:conf/aaai/SerbanSBCP16}. % is an extension of the \emph{HRED} model (hierarchical recurrent encoder-decoder model) 
The model decomposes dialogues using a two-level hierarchy: sequences of utterances (e.g.\@ sentences), and sub-sequences of tokens (e.g.\@ words).
Let $\mathbf{w}_n$ be the $n$'th utterance in a dialogue with $N$ utterances.
Let $w_{n, m}$ be the $m$'th word in the $n$'th utterance from vocabulary $V$ given as a 1-of-$|V|$ binary encoding. Let $M_n$ be the number of words in the $n$'th utterance.
For each utterance $n=1,\dots,N$, the model generates a latent variable $z_n$.
Conditioned on this latent variable, the model then generates the next utterance:
%\begin{align}
%& P_\theta(\mathbf{w}_1, \ldots, \mathbf{w}_N, z_n) \nonumber \\ &= \prod_{n = 1}^N P_\theta(\mathbf{w}_n | \mathbf{w}_{< n}, z_n) P_{\theta}(z_n | \mathbf{w}_{< n}) \nonumber \\
%& = \prod_{n = 1}^N \prod_{m = 1}^{M_n} P_\theta(w_{n, m} | w_{n, < m}, \mathbf{w}_{< n}, z_n) P_{\theta}(z_n | \mathbf{w}_{< n}),
%\end{align}
\begin{align}
P_\theta(\mathbf{w}_1, z_1, \ldots, \mathbf{w}_N, z_N) = \prod_{n = 1}^N P_{\theta}(z_n | \mathbf{w}_{< n}) & \nonumber \\ % \prod_{n = 1}^N P_\theta(\mathbf{w}_n | \mathbf{w}_{< n}, z_n) P_{\theta}(z_n | \mathbf{w}_{< n}) \nonumber \\
 \times \prod_{m = 1}^{M_n} P_\theta(w_{n, m} | w_{n, < m}, \mathbf{w}_{< n}, z_n), & \nonumber
\end{align}
where $\theta$ are the model parameters.
\emph{VHRED} consists of three RNN modules: an \textit{encoder} RNN, a \textit{context} RNN and a \textit{decoder} RNN.
The \textit{encoder} RNN computes an embedding for each utterance.
This embedding is fed into the \textit{context} RNN, which computes a hidden state summarizing the dialogue context before utterance $n$: $h^{\text{con}}_{n-1}$.
This state represents the additional conditioning information, which is used to compute the prior distribution over $z_n$:
\begin{align}
P_{\theta}(z_n \mid 
\mathbf{w}_{<n}) = f^{\text{prior}}_\theta (z_n ; h_{n-1}^{con}), \nonumber
\end{align}
where $f^{\text{prior}}$ is a PDF parametrized by both $\theta$ and $h_{n-1}^{\text{con}}$.
A sample is drawn from this distribution: $z_n \sim P_\theta (z_n | \mathbf{w}_{<n})$.
This sample is given as input to the \textit{decoder} RNN, which then computes the output probabilities of the words in the next utterance.
The model is trained by maximizing the variational lower-bound, which factorizes into independent terms for each sub-sequence (utterance):
\begin{align}
\log P_{\theta} & (\mathbf{w}_1, \dots, \mathbf{w}_N) \nonumber \\ \geq  \sum_{n=1}^N & - \text{KL} \left [ Q_{\psi}(z_n \mid \mathbf{w}_1, \dots, \mathbf{w}_n) || P_{\theta}(z_n \mid \mathbf{w}_{<n} ) \right ] \nonumber \\
& + \mathbb{E}_{Q_{\psi}(z_n \mid \mathbf{w}_1, \dots, \mathbf{w}_n)} \left [ \log P_{\theta}(\mathbf{w}_n \mid z_n, \mathbf{w}_{< n}) \right ], \nonumber % \label{VHRED:lower_bound}
\end{align}
where distribution $Q_{\psi}$ is the approximate posterior distribution with parameters $\psi$, computed similarly as the prior distribution but further conditioned on the \textit{encoder} RNN hidden state of the next utterance.
% Aaron: We need more detail following equation 10 for the hybrid model. Make the factoring into the piecewise and Gaussian terms explicit.

%\begin{align}
%Q_{\psi}(z_n \mid \mathbf{w}_{\leq n}) = f^{\text{post}}_\psi (h_{n-1}^{con}, h_{n, M_n}^{enc}),
%\end{align}
%where $f^{\text{post}}$ is a PDF.

The original \emph{VHRED} model \citep{serban2016hierarchical} used Gaussian latent variables.
We refer to this model as \emph{G-VHRED}.
The first model we propose uses piecewise constant latent variables instead of Gaussian latent variables.
We refer to this model as \emph{P-VHRED}.
The second model we propose takes advantage of the representation power of both Gaussian and piecewise constant latent variables.
This model samples both a Gaussian latent variable $z_n^{\text{gaussian}}$ and a piecewise latent variable $z_n^{\text{piecewise}}$ independently conditioned on the \textit{context} RNN hidden state:
\begin{align}
P_{\theta}(z_n^{\text{gaussian}} \mid 
\mathbf{w}_{<n}) &= f^{\text{prior, gaussian}}_\theta (z_n^{\text{gaussian}} ; h_{n-1}^{con}), \nonumber \\
P_{\theta}(z_n^{\text{piecewise}} \mid 
\mathbf{w}_{<n}) &= f^{\text{prior, piecewise}}_\theta (z_n^{\text{piecewise}} ; h_{n-1}^{con}), \nonumber
\end{align}
where $f^{\text{prior, gaussian}}$ and $f^{\text{prior, piecewise}}$ are PDFs parametrized by independent subsets of parameters $\theta$.
We refer to this model as \emph{H-VHRED}.

\section{Experiments}
\label{experiments}
We evaluate the proposed models on two types of natural language processing tasks: document modeling and dialogue natural language generation.
%In order to evaluate the  the ability of our piecewise latent variables to capture complex aspects of data distributions, we conduct experiments with both the NVDM and VHRED models.
All models are trained with back-propagation using the variational lower-bound on the log-likelihood or the exact log-likelihood. We use the first-order gradient descent optimizer Adam \citep{kingma2014adampublished} with gradient clipping \citep{pascanu2012difficulty}\footnote{Code and scripts are available at \url{https://github.com/ago109/piecewise-nvdm-emnlp-2017} and \url{https://github.com/julianser/hred-latent-piecewise}.} % , where hyper-parameter choices varied depending on the task.
% Alex: now that we both used Adam, I've moved it back here and updated each sub-section accordingly =]
%The specifics of the design of the encoder and decoder differed between the two tasks, as described in Sections \ref{nvdm} and \ref{vhred}. 

\subsection{Document Modeling}
\label{doc_model}
% Furthermore, \textit{G-NVDM} outperform the previous state-of-the-art on 20-NG.
\begin{table}[t]
  \caption{Test perplexities on three document modeling tasks: 20-NewGroup (20-NG), Reuters corpus (RCV1) and CADE12 (CADE). Perplexities were calculated using 10 samples to estimate the variational lower-bound. The \textit{H-NVDM} models perform best across all three datasets.} \label{tabel:document_results}
  %\small
  \centering
    \begin{tabular}{lccccccccccccc}
    %\toprule
    \textbf{Model} & \textbf{20-NG} & \textbf{RCV1} & \textbf{CADE} \\
    \midrule
    \textit{LDA} & $1058$ & $--$ & $--$ \\
%    \textit{RSM} & $953$ & $--$ & $--$ \\
    \textit{docNADE} & $896$ & $--$ & $--$ \\
%    \textit{SBN} & $909$ & $--$ & $--$ \\
%    \textit{fDARN} & $917$ & $--$ & $--$ \\
    \textit{NVDM} & $836$ & $--$ & $--$ \\ \midrule
    \textit{G-NVDM} & $651$ & $905$ & $339$ \\
    \textit{H-NVDM-3} & $607$ & $865$ & $\mathbf{258}$ \\
    \textit{H-NVDM-5} & $\mathbf{566}$ & $\mathbf{833}$ & $294$ \\ \bottomrule
    \end{tabular}
\end{table}

\textbf{Tasks} We use three different datasets for document modeling experiments.
First, we use the 20 News-Groups (20-NG) dataset \citep{hinton_softmax_2009}.
Second, we use the Reuters corpus (RCV1-V2), using a version that contained a selected 5,000 term vocabulary. %\footnote{Scripts and code will be made available upon publication.}%\footnote{We will make the code and scripts used to create the final document input vectors and vocabulary files publicly available upon publication.}
As in previous work \citep{hinton_softmax_2009,larochelle_neural_2012}, we transform the original word frequencies using the equation $\log(1 + \text{TF})$, where TF is the original word frequency.
Third, to test our document models on text from a non-English language, we use the Brazilian Portuguese CADE12 dataset \citep{2007:phd-Ana-Cardoso-Cachopo}.
%We pre-processed the dataset by stemming words and removing stop words. We further filtered terms that occurred less than 130 times to obtain a vocabulary of 3,736 terms (over 26,991 training and 13,486 test documents).
For all datasets, we track the validation bound on a subset of 100 vectors randomly drawn from each training corpus.

\begin{table}[!t]
%% increase table row spacing, adjust to taste
%\renewcommand{\arraystretch}{1.4}do
\caption{Word query similarity test on 20 News-Groups: for the query `space'', we retrieve the top $10$ nearest words in word embedding space based on Euclidean distance. \textit{H-NVDM-5} associates multiple meanings to the query, while \textit{G-NVDM} only associates the most frequent meaning.} %more general/abstract terms to the query, which may or may not always be what is desired.}
\label{word_query}
\centering
\begin{tabular}{lllll}
%\hline\hline
\multicolumn{1}{l}{\begin{tabular}[x]{@{}c@{}}\textbf{G-NVDM}\\\end{tabular}}&\multicolumn{1}{c}{\begin{tabular}[x]{@{}c@{}}\textbf{H-NVDM-3}\\\end{tabular}}&\multicolumn{1}{c}{\begin{tabular}[x]{@{}c@{}}\textbf{H-NVDM-5}\\\end{tabular}}\tabularnewline
\hline
environment & project & science \tabularnewline
project & gov & built \tabularnewline
flight & major & high \tabularnewline
lab & based & technology \tabularnewline
mission & earth & world \tabularnewline
launch & include & form \tabularnewline
field & science & scale \tabularnewline
working & nasa & sun \tabularnewline
build & systems & special \tabularnewline
gov & technical & area \tabularnewline
\hline
\end{tabular}
\end{table}

%In addition, we make use of the NIPS XXX collection, which has also been pre-processed and publicly available.\footnote{http://www.cs.nyu.edu/~roweis/data.html}. The data-set contains XXXX documents with a vocabulary of XXXX...

\textbf{Training} All models were trained using mini-batches with 100 examples each. % over 150 training set epochs.
A learning rate of $0.002$ was used.
Model selection and early stopping were conducted using the validation lower-bound, estimated using five stochastic samples per validation example.
Inference networks used 100 units in each hidden layer for 20-NG and CADE, and 100 for RCV1.
We experimented with both $50$ and $100$ latent random variables for each class of models, and found that $50$ latent variables performed best on the validation set.
For \emph{H-NVDM} we vary the number of components used in the PDF, investigating the effect that 3 and 5 pieces had on the final quality of the model.
The number of hidden units was chosen via preliminary experimentation with smaller models.
On 20-NG, we use the same set-up as \citep{hinton_softmax_2009} and therefore report the perplexities of a topic model (\emph{LDA}, \citep{hinton_softmax_2009}), 
%the Replicated Softmax (\emph{RSM}, \citep{hinton_softmax_2009}),
the document neural auto-regressive estimator (\emph{docNADE}, \citep{larochelle_neural_2012}),
%a sigmoid belief network (\emph{SBN}, \citep{mnih2014neural}), 
%a deep auto-regressive neural network (\emph{fDARN}, \citep{mnih2014neural}),
and a neural variational document model with a fixed standard Gaussian prior (\emph{NVDM}, lowest reported perplexity, \citep{miao2015neural}).

\textbf{Results} In Table \ref{tabel:document_results}, we report the test document perplexity: $\exp(-\frac{1}{D}\sum_{n}\frac{1}{L_n} \log P_{\theta}(x_n)$. 
We use the variational lower-bound as an approximation based on 10 samples, as was done in \citep{mnih2014neural}. 
First, we note that the best baseline model (i.e.\@ the \emph{NVDM}) is more competitive when both the prior and posterior models are learnt together (i.e.\@ the \emph{G-NVDM}), as opposed to the fixed prior of \citep{miao2015neural}.
Next, we observe that integrating our proposed piecewise variables yields even better results in our document modeling experiments, substantially improving over the baselines. More importantly, in the 20-NG and Reuters datasets, increasing the number of pieces from 3 to 5 further reduces perplexity. 
Thus, we have achieved a new state-of-the-art perplexity on 20 News-Groups task and --- to the best of our knowledge -- better perplexities on the CADE12 and RCV1 tasks compared to using a state-of-the-art model like the \emph{G-NVDM}.
%Furthermore, we found that using iterative inference yielded yet a further boost in performance, estimating a tighter bound (see Appendix).
We also evaluated the converged models using an non-parametric inference procedure, where a separate approximate posterior is learned for each test example in order to tighten the variational lower-bound.
\textit{H-NVDM} also performed best in this evaluation across all three datasets, which confirms that the performance improvement is due to the piecewise components.
See appendix for details.
%The second column \emph{SGD-Inf}, refers to the model's test-perplexity when the lower-bound is tightened using iterative inference to search for the optimal latent variable per document.
%See appendix for further details on the iterative inference procedure.

%, however, this form of inference is expensive and requires additional meta-parameters (e.g.\@ a step-size and an early-stopping criterion). %We remark a simpler, and more accurate, approach to inference would be to use importance sampling. 
%In some cases, 3 pieces appear to be sufficient, although this might simply indicate the need for yet further optimization of the piecewise variables, as hinted by the Reuters models (where iterative inference actually helped the 5-piece model achieve lowest perplexity). <-- since I redid the experiment, I was able to optimize the 5-piece model enough, it ended up performing the best in the end

\begin{figure}[!t]
\centering
\includegraphics[scale=0.20]{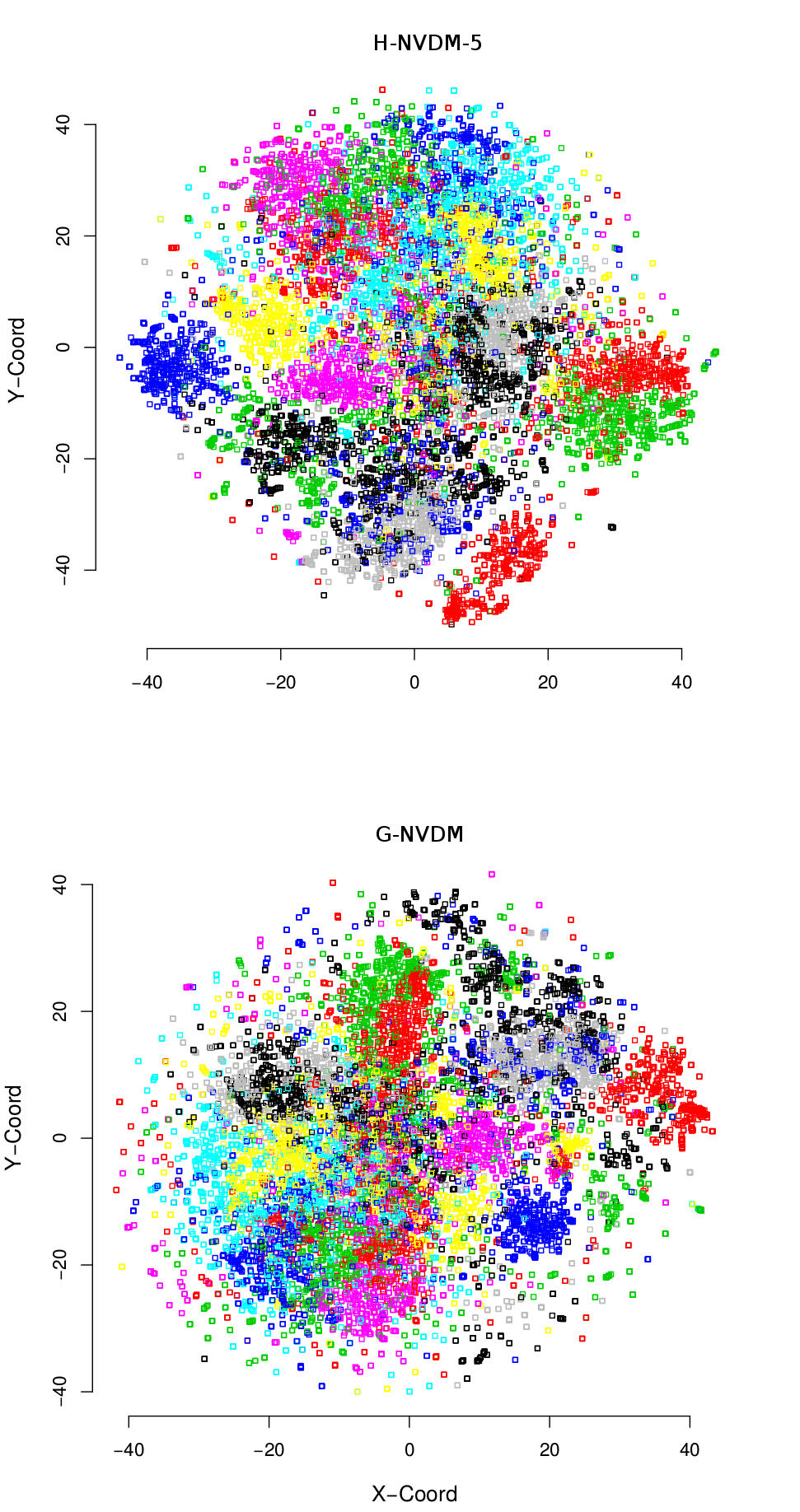}
\caption{Latent variable approximate posterior means t-SNE visualization on 20-NG for \textit{G-NVDM} and \textit{H-NVDM-5}. Colors correspond to the topic labels assigned to each document.}
\label{fig:tsne}
\end{figure}

In Table \ref{word_query}, we examine the top ten highest ranked words given the query term ``space'', using the decoder parameter matrix. %(since the decoder is directly affected by the latent variables in our document models).
The piecewise variables appear to have a significant effect on what is uncovered by the model.%, as each model returns different results for the query word.
In the case of ``space'', the hybrid with 5 pieces seems to value two senses of the word--one related to ``outer space'' (e.g., ``sun'', ``world'', etc.) and another related to the dimensions of depth, height, and width within which things may exist and move (e.g., ``area'', ``form'', ``scale'', etc.).
On the other hand, \textit{G-NVDM} appears to only capture the ``outer space'' sense of the word.
More examples are in the appendix.

Finally, we visualized the means of the approximate posterior latent variables on 20-NG through a t-SNE projection. As shown in Figure \ref{fig:tsne}, both \textit{G-NVDM} and \textit{H-NVDM-5} learn representations which disentangle the topic clusters on 20-NG. However, \textit{G-NVDM} appears to have more dispersed clusters and more outliers (i.e.\@ data points in the periphery) compared to \textit{H-NVDM-5}.
Although it is difficult to draw conclusions based on these plots, these findings could potentially be explained by the Gaussian latent variables fitting the latent factors poorly.

%In our current examples, it appears that the \emph{H-NVDM} with 5 pieces returns more general words. For example, in the case of ``government'', the baseline seems to value the plural form of the word (which is largely based on morphology) while the hybrid model actually pulls out meaningful terms such as ``federal'', ``policy'', and ``administration''. 

%We also qualitatively evaluate the quality of the latent document representations learned by the various document models through a t-SNE visualization \citep{maaten2008visualizing}, specifically employing the Barnes-Hutt approximation for scalability. As observed in Figure XXXX, it appears that...

\subsection{Dialogue Modeling}
\label{dialog_model}

% \begin{table}[t]
%   \caption{Ubuntu evaluation using precision (P), recall (R) and F1 metrics w.r.t.\@ activities and entities. \textit{G-VHRED}, \textit{P-VHRED} and \textit{H-VHRED} all outperform the baseline \textit{HRED}. \textit{G-VHRED} performs best w.r.t.\@ activities and \textit{H-VHRED} performs best w.r.t.\@ entities.} \label{tabel:ubuntu_results}
%   \small
%   \centering
%     \begin{tabular}{lccccccccccccc}
%     %\toprule
%      & \multicolumn{3}{c}{\textbf{Activity}} & \multicolumn{3}{c}{\textbf{Entity}} \\ \midrule
%     \textbf{Model} & \textbf{P\@} & \textbf{R\@} & \textbf{F1} & \textbf{P\@} & \textbf{R\@} & \textbf{F1} \\
%     \midrule
%     \textit{HRED} & $6.55$ & $4.49$ & $4.77$ & $3.09$ & $2.3$ & $2.43$ \\
%     \textit{G-VHRED} & $10.89$ & $\mathbf{10.11}$ & $\mathbf{9.24}$ & $3.11$ & $2.44$ & $2.49$ \\
%     \textit{P-VHRED} & $6.79$ & $4.67$ & $5$ & $3.08$ & $2.4$ & $2.49$ \\   
%     \textit{H-VHRED} & $\mathbf{10.92}$ & $8.31$ & $8.41$ & $\mathbf{4.81}$	& $\mathbf{3.51}$ & $\mathbf{3.72}$ \\ \bottomrule
%     \end{tabular}
% \end{table}

\begin{table}[t]
  \caption{Ubuntu evaluation using F1 metrics w.r.t.\@ activities and entities. \textit{G-VHRED}, \textit{P-VHRED} and \textit{H-VHRED} all outperform the baseline \textit{HRED}. \textit{G-VHRED} performs best w.r.t.\@ activities and \textit{H-VHRED} performs best w.r.t.\@ entities.} \label{tabel:ubuntu_results}
  %\small
  \centering
    \begin{tabular}{lccccccccccccc}
    %\toprule
    \textbf{Model} & \textbf{Activity} & \textbf{Entity} \\
    \midrule
    \textit{HRED} & $4.77$ & $2.43$ \\
    \textit{G-VHRED} & $\mathbf{9.24}$ & $2.49$ \\
    \textit{P-VHRED} & $5$ & $2.49$ \\   
    \textit{H-VHRED} & $8.41$ & $\mathbf{3.72}$ \\ \bottomrule
    \end{tabular}
\end{table}

\textbf{Task} We evaluate \emph{VHRED} on a natural language generation task, where the goal is to generate responses in a dialogue.
This is a difficult problem, which has been extensively studied in the recent literature \citep{ritter2011data,lowe2015ubuntu,sordoni2015aneural,li2015diversity,DBLP:conf/aaai/SerbanSBCP16,serban2016generative}.
Dialogue response generation has recently gained a significant amount of attention from industry, with high-profile projects such as Google SmartReply \citep{kannan2016smart} and Microsoft Xiaoice \citep{markoff2015forsymp}.
Even more recently, Amazon has announced the Alexa Prize Challenge for the research community with the goal of developing a natural and engaging chatbot system \citep{farber2015amazon}.

We evaluate on the technical support response generation task for the Ubuntu operating system.
We use the well-known Ubuntu Dialogue Corpus \citep{lowe2015ubuntu,lowe2017ubuntu}, which consists of about 1/2 million natural language dialogues extracted from the \#Ubuntu Internet Relayed Chat (IRC) channel.
The technical problems discussed span a wide range of software-related and hardware-related issues.
Given a dialogue history --- such as a conversation between a user and a technical support assistant --- the model must generate the next appropriate response in the dialogue.
For example, when it is the turn of the technical support assistant, the model must generate an appropriate response helping the user resolve their problem.

We evaluate the models using the activity- and entity-based metrics designed specifically for the Ubuntu domain \citep{serban2017multires}.
These metrics compare the \textit{activities} and \textit{entities} in the model generated responses with those of the reference responses; activities are verbs referring to high-level actions (e.g.\@ \textit{download}, \textit{install}, \textit{unzip}) and entities are nouns referring to technical objects (e.g.\@ \textit{Firefox}, \textit{GNOME}).
The more activities and entities a model response overlaps with the reference response (e.g.\@ expert response) the more likely the response will lead to a solution.%~\citep{serban2017multires}.

\textbf{Training} The models were trained to maximize the log-likelihood of training examples using a learning rate of $0.0002$ and mini-batches of size $80$.
We use a variant of truncated back-propagation.
%and apply gradient clipping \citep{pascanu2012difficulty}.
We terminate the training procedure for each model using early stopping, estimated using one stochastic sample per validation example.
We evaluate the models by generating dialogue responses: conditioned on a dialogue context, we fix the model latent variables to their median values and then generate the response using a beam search with size 5.
We select model hyper-parameters based on the validation set using the F1 activity metric, as described earlier.
%, of the generated responses on the validation set.

It is often difficult to train generative models for language with stochastic latent variables \citep{bowman2015generating,serban2016hierarchical}.
For the latent variable models, we therefore experiment with reweighing the KL divergence terms in the variational lower-bound with values $0.25$, $0.50$, $0.75$ and $1.0$.
In addition to this, we linearly increase the KL divergence weights starting from zero to their final value over the first $75000$ training batches.
Finally, we weaken the \textit{decoder} RNN by randomly replacing words inputted to the decoder RNN with the unknown token with $25\%$ probability.
These steps are important for effectively training the models, and the latter two have been used in previous work by \citet{bowman2015generating} and \citet{serban2016hierarchical}.
%At test time, we use beam search with $5$ beams for outputting responses with the RNN decoders.
%At the beginning of each beam search, samples of the latent variables are generated and conditioned on throughout the beam search. We fix the word embedding dimensionality to $400$.

{\bf HRED (Baseline): }
We compare to the \emph{HRED} model \citep{DBLP:conf/aaai/SerbanSBCP16}: a sequence-to-sequence model, shown to outperform other established models on this task, such as the LSTM RNN language model \citep{serban2017multires}.
The \emph{HRED} model's \textit{encoder} RNN uses a bidirectional GRU RNN encoder, where the forward and backward RNNs each have $1000$ hidden units.
The context RNN is a GRU encoder with $1000$ hidden units, and the decoder RNN is an LSTM decoder with $2000$ hidden units.\footnote{Since training lasted between 1-3 weeks for each model, we had to fix the number of hidden units during preliminary experiments on the training and validation datasets.}
The encoder and context RNNs both use layer normalization \citep{ba2016layer}.\footnote{We did not apply layer normalization to the decoder RNN, because several of our colleagues have found that this may hurt the performance of generative language models.}
We also experiment with an additional rectified linear layer applied on the inputs to the decoder RNN.
As with other hyper-parameters, we choose whether to include this additional layer based on the validation set performance.
\emph{HRED}, as well as all other models, use a word embedding dimensionality of size $400$.

{\bf G-HRED: }
We compare  to \emph{G-VHRED}, which is \emph{VHRED} with Gaussian latent variables \citep{serban2016hierarchical}.
\emph{G-VHRED} uses the same hyper-parameters for the encoder, context and decoder RNNs as the HRED model. 
The model has $100$ Gaussian latent variables per utterance.

{\bf P-HRED: }
The first model we propose is \emph{P-VHRED}, which is \emph{VHRED} model with piecewise constant latent variables.
We use $n=3$ number of pieces for each latent variable.
\emph{P-VHRED} also uses the same hyper parameters for the encoder, context and decoder RNNs as the \emph{HRED} model. 
Similar to \emph{G-VHRED}, \emph{P-VHRED} has $100$ piecewise constant latent variables per utterance.

{\bf H-HRED: }
The second model we propose is \emph{H-VHRED}, which has $100$ piecewise constant (with $n=3$ pieces per variable) and $100$ Gaussian latent variables per utterance.
\emph{H-VHRED} also uses the same hyper-parameters for the encoder, context and decoder RNNs as \emph{HRED}.

{\bf Results: }
The results are given in Table \ref{tabel:ubuntu_results}.
All latent variable models outperform \emph{HRED} w.r.t.\@ both activities and entities.
This strongly suggests that the high-level concepts represented by the latent variables help generate meaningful, goal-directed responses.
Furthermore, each type of latent variable appears to help with a different aspects of the generation task.
\emph{G-VHRED} performs best w.r.t.\@ activities (e.g.\@ \textit{download}, \textit{install} and so on), which occur frequently in the dataset.
This suggests that the Gaussian latent variables learn useful latent representations for frequent actions.
On the other hand, \emph{H-VHRED} performs best w.r.t.\@ entities (e.g.\@ \textit{Firefox}, \textit{GNOME}), which are often much rarer and mutually exclusive in the dataset.
This suggests that the combination of Gaussian and piecewise latent variables help learn useful representations for entities, which could not be learned by Gaussian latent variables alone.
%This suggests that the Gaussian latent variables learn useful latent representations for frequent actions.
%On the other hand, \emph{H-VHRED} performs best w.r.t.\@ entities, which are rarer than the activities and often mutually exclusive.
%This suggests a combination of Gaussian latent variables and piecewise constant latent variables are necessary to represent the large and diverse set of objects in the Ubuntu domain.
We further conducted a qualitative analysis of the model responses, which supports these conclusions. See Appendix \ref{appendix_f}.\footnote{Results on a Twitter dataset are given in the appendix.}

%We also compare to the HRED model \citep{DBLP:conf/aaai/SerbanSBCP16}.
%The HRED model \textit{encoder} RNN has a bidirectional GRU RNN encoder, where the forward and backward RNNs each have $1000$ hidden units.
%The encoder and context RNNs use layer normalization \citep{ba2016layer}.
%We experiment with $500$ and $1000$ hidden units for the context RNN and with $1000$ and $2000$ hidden units for the LSTM RNN decoder.
%Based on the validation log-likelihood,
%we choose $1000$ hidden units for both the context RNN and decoder RNN.

\section{Conclusions}
\label{conc}
In this paper, we have sought to learn rich and flexible multi-modal representations of latent variables for complex natural language processing tasks.
We have proposed the piecewise constant distribution for the variational autoencoder framework.
We have derived closed-form expressions for the necessary quantities required for in the autoencoder framework,
and proposed an efficient, differentiable implementation of it.
%We have asserted the theoretical power and flexibility of the proposed piecewise constant distribution, and derived closed-form expressions for the necessary quantities required for in the autoencoder framework.
%developed the piecewise constant prior, which can be efficiently and flexibly adjusted to capture distributions with complex probability manifolds spanning many modes, such as those over topics.
We have incorporated the proposed piecewise constant distribution into two model classes --- \emph{NVDM} and \emph{VHRED} --- and evaluated the proposed models on document modeling and dialogue modeling tasks.
We have achieved state-of-the-art results on three document modeling tasks, and have demonstrated substantial improvements on a dialogue modeling task.
Overall, the results highlight the benefits of incorporating the flexible, multi-modal piecewise constant distribution into variational autoencoders.
%have shown the effectiveness of our framework in building models capable of learning richer structure from data.
%In particular, we have demonstrated new state-of-the-art results on several document modeling tasks.
Future work should explore other natural language processing tasks, where the data is likely to arise from complex, multi-modal latent factors.
%Two potential avenues such as online debates \citep{rosenthal2015couldn},
%as well as tasks where additional information is available, such as in semi-supervised document categorization \citep{ororbia2015learning}.
%Furthermore, the piecewise variables proposed in this work could prove useful in uncovering interesting and novel information in lesser-explored corpora.

\section*{Acknowledgments}
The authors acknowledge NSERC,
Canada  Research  Chairs, CIFAR, IBM Research, Nuance Foundation and Microsoft Maluuba for funding.
Alexander G. Ororbia II was funded by a NACME-Sloan scholarship.
The authors thank Hugo Larochelle for sharing the NewsGroup 20 dataset.
The authors thank Laurent Charlin, Sungjin Ahn, and Ryan Lowe for constructive feedback.
This  research  was enabled  in  part  by  support  provided  by  Calcul Qubec (\url{www.calculquebec.ca}) and Compute Canada (\url{www.computecanada.ca}).
\newpage
%\nocitep{langley00}

%\bibliography{ref}
%\bibliographystyle{icml2017}

\bibliography{ref}
\bibliographystyle{emnlp_natbib}

\newpage
\appendix

%\newpage
%\ 
\newpage

\section{Appendix: Inappropriate Gaussian Priors} \label{appendix_0}
The majority of work on VAEs propose to parametrize $z$ --- both the prior and approximate posterior (encoder) --- as a multivariate Gaussian variable.
However, the multivariate Gaussian is a uni-modal distribution and can therefore only represent one mode in latent space.
Furthermore, the multivariate Gaussian is perfectly symmetric with a constant kurtosis. % Furthermore, the multivariate Gaussian is perfectly symmetric and has a fixed kurtosis. % 
% Aaron: This just isn't true for high dim. Gaussians. In fact, as the dim of z increases the probability mass close to the mode (let's say within 1 std-dev.) goes to zero (see e.g. http://math.stackexchange.com/questions/143377/3-sigma-rule-for-multivariate-normal-distribution).
These properties are problematic if the latent variables we aim to represent are inherently multi-modal, or if the latent variables follow complex, non-linear probability manifolds (e.g.\@ asymmetric distributions or heavy-tailed distributions).
For example. the frequency of topics in news articles could be represented by a continuous probability distribution, where each topic has its own island of probability mass; \textit{sports} and \textit{politics} topics might each be clustered on their own separate island of probability mass with zero or little mass in between them.
Due to its uni-modal nature, the Gaussian distribution can never represent such probability distributions.
As another example, ambiguity and uncertainty in natural language conversations could similarly be represented by islands of probability mass; given the question \textit{How do I install Ubuntu on my laptop?}, a model might assign positive probability mass to specific, unambiguous entities like \textit{Ubuntu 4.10} and to well-defined procedures like \textit{installation using a DVD}. % or \textit{installation using a USB key}.
In particular, certain entities like \textit{Ubuntu 4.10} are now outdated --- these entities occur rarely in practice and should be considered rare events.
When modeling such complex, multi-modal latent distributions, the mapping from multivariate Gaussian latent variables to outputs --- i.e.\@ the conditional distribution $P_{\theta}(w_n | z)$ --- has to be highly non-linear in order to compensate for the simplistic Gaussian distribution and capture the natural latent factors in an intermediate layer of the model.
However, it is difficult to learn such non-linear mappings 
%with existing stochastic optimization methods.
%such as mini-batch stochastic gradient descent and its variants.
%Learning is particularly difficult 
when using the variational bound in eq.\@ \eqref{eq:variatonal_lower_bound}, as it incurs additional variance from sampling the latent variable $z$.
Consequently, such models are likely to converge on solutions that do not capture salient aspects of the latent variables, which in turn leads to a poor fit of the output distribution.
%Consequently, such a model is very likely to converge on a solution which does not model multi-modality which then leads to a poor approximation of the output distribution.
% FOOTNOTE
%\footnote{Some work has in fact used a mixture of Gaussians as posterior approximation, but due to tractability, the number of mixture components (modes) is typically very small.}

\section{Appendix: Piecewise Constant Variable Derivations}
\label{appendix_a}

To train the model using the re-parametrization trick, we need to generate $z = f(\epsilon)$ where $\epsilon \sim \text{Uniform}(0, 1)$. To do so, we employ inverse transform sampling \citep{devroye1986sample},
%\url{https://en.wikipedia.org/wiki/Inverse_transform_sampling}
which requires finding the inverse of the cumulative distribution function (CDF).
We derive the CDF of eq.\@ \eqref{piecewise_pdf}:
\begin{align}
\phi(z) = \dfrac{1}{K} \sum_{i=1}^n & 1_{ \left (\dfrac{i}{n} \leq z \right )} K_i + 1_{ \left (\dfrac{i-1}{n} \leq z \leq \dfrac{i}{n} \right )} \nonumber \\ & * \left (z - \dfrac{i-1}{n} \right ) a_i. 
\end{align}
Next, we derive its inverse:
\begin{align}
\phi^{-1}(\epsilon) = \sum_{i=1}^n & 1_{ \left ( \dfrac{1}{K} \sum_{j=0}^{i-1} K_j \leq \epsilon \leq \dfrac{1}{K} \sum_{j=0}^{i} K_j  \right )} \nonumber \\ & * \left ( \dfrac{i-1}{n} + \dfrac{K}{a_i} \left ( \epsilon - \dfrac{1}{K} \sum_{j=0}^{i-1} K_j \right ) \right ) %\label{piece_inv_cdf}
\end{align}
Armed with the inverse CDF, we can now draw a sample $z$:
\begin{align}
z = \phi^{-1}(\epsilon), \quad \text{where} \ \epsilon \sim \text{Uniform}(0, 1).
\end{align}
In addition to sampling, we need to compute the Kullback-Leibler (KL) divergence between the prior and approximate posterior distributions of the piecewise constant variables.
We assume both the prior and the posterior are piecewise constant distributions.
We use the \textit{prior} superscript to denote prior parameters and the \textit{post} superscript to denote posterior parameters (encoder model parameters).
The KL divergence between the prior and posterior can be computed using a sum of integrals, where each integral inside the sum corresponds to one constant segment:
{\fontsize{9.75}{11.7}
\begin{align}
& \text{KL} \left [ Q_{\psi}(z | w_1, \dots, w_N) || P_{\theta}(z) \right ] \nonumber \\ & = \int_0^1 Q_{\psi}(z | w_1, \dots, w_N) \log \left ( \dfrac{Q_{\psi}(z|  w_1, \dots, w_N)}{P_{\theta}(z)}\right ) dz
 \\ & = \sum_{i=1}^n \int_{0}^{1/n} \dfrac{a_i^{\text{post}}}{K^{\text{post}}} \log \left ( \dfrac{a_i^{\text{post}}/K^{\text{post}}}{a_i^{\text{prior}}/K^{\text{prior}}} \right ) dz \\ 
 & = \dfrac{1}{n} \sum_{i=1}^n \dfrac{a_i^{\text{post}}}{K^{\text{post}}}
 \log \left ( \dfrac{a_i^{\text{post}}/K^{\text{post}}}{a_i^{\text{prior}}/K^{\text{prior}}} \right )
 \\ & = \dfrac{1}{n} \dfrac{1}{K^{\text{post}}} \sum_{i=1}^n a_i^{\text{post}} \left ( \log (a_i^{\text{post}}) - \log(a_i^{\text{prior}}) \right )  \nonumber \\
 & \quad + \log(K^{\text{prior}}) - \log(K^{\text{post}})
\end{align}
}%

%In order to train the model, we take partial derivatives of the variational bound in eq.\@ \eqref{eq:variatonal_lower_bound} w.r.t.\@ each parameter in $\theta$ and $\psi$.
%These expressions involve derivatives of the indicator functions, which have derivatives zero everywhere except for the changing points where the derivative is undefined.
%However, the probability of sampling $\epsilon$ such that an indicator function is exactly at its changing point is effectively zero.
%Therefore, we fix their derivatives to zero.\footnote{We thank Christian A.\@ Naesseth for pointing out this assumption.}
%A similar approach is used for training neural networks with rectified linear units.

In order to improve training, we further transform the piecewise constant latent variables to lie within the interval $[-1, 1$] after sampling: $z' = 2 z - 1$. This ensures the input to the decoder RNN has mean zero initially.% at the beginning of training.

%in order to integrate eq.\@  \eqref{piece_inv_cdf} and \eqref{kl_piece}.
%into an automatic differentiation framework (the derivations of which can be found in our publicly available code at XXXX).

\section{Appendix: NVDM Implementation}
\label{appendix_c}
The complete \textit{NVDM} architecture is defined as:
\begin{align*}
\pi(W) &= f^0(E^0 W + b^0), \\
Enc(W) &= f^1(E^1 \pi(W) + b^1), \\
z_{Gaussian} &= \mu^{\text{post}} + \sqrt{ \sigma^{2,\text{post}} } \otimes \epsilon_0, \\
z_{Piecewise} &= \phi^{-1,post}(\epsilon_1),\\
z &= \langle z_{Gaussian}, z_{Piecewise} \rangle,\\
Dec(w, z) &= g(-w^{\text{T}} R z),
\end{align*}
%Dec(w, z) &= g((E^0 w)^{\text{T}} R z),
where $\otimes$ is the Hadamard product, $\langle \circ, \circ \rangle$ is an operator that combines the Gaussian and the Piecewise variables and $Dec(w, z)$ is the decoder model.\footnote{Operations include vector concatenation, summation, or averaging.}
As a result of using the re-parametrization trick and choice of prior, we calculate the latent variable $z$ through the two samples, $\epsilon_0$ and $\epsilon_1$.
$f(\circ)$ is a non-linear activation function, which was the parametrized linear rectifier (with a learnable ``leak'' parameters) for the 20 News-Groups experiments and the softsign function, or $f(v) = v / (1 + |v|)$, for Reuters and CADE.
The decoder model $Dec(z)$ outputs a probability distribution over words conditioned on $z$. 
In this case, we define $g(\circ)$ as the softmax function (omitting the bias term $c$ for clarity) computed as:
\begin{align*}
Dec(w, z) = P_{\theta}(w|z) = \frac{\exp{( -w^{\text{T}} R z )}}{\sum_{w'} \exp{(-w^{\text{T}} R z)}}, \label{sample_y_given_h1_h2}
\end{align*}
% Alex: I added a negative sign in my decoder during my debugging to more closely match Maio et al paper since I ended up using that: smx(-R h + b)
The decoder's output is used to calculate the first term in the variational lower-bound: $\log P_{\theta}(W|z)$. The prior and posterior distributions are used to compute the KL term in the variational lower-bound.
The lower-bound is:
\begin{align*}
\mathcal{L} = & \text{E}_{Q_{\psi}(z | W) } \Bigg[ \sum^{N}_{i=1} \log P_{\theta}(w_i | z) \Bigg] \nonumber \\ & - \text{KL} \left [ Q_{\psi}(z | W) || P_{\theta}(z) \right ],
\end{align*}
where the KL term is the sum of the Gaussian and piecewise KL-divergence measures:
\begin{align*}
 \text{KL} & \left [ Q(z | W) || P(z) \right ] \\ = & \text{KL}_{Gaussian} \left [ Q(z | W) || P(z) \right ] \\ & + \text{KL}_{Piecewise} \left [ Q(z | W) || P(z) \right ].
\end{align*}
The KL-terms may be interpreted as regularizers of the parameter updates for the encoder model \citep{kingma2013auto}. These terms encourage the posterior distributions to be similar to their corresponding prior distributions, by limiting the amount of information the encoder model transmits regarding the output.
%For example, it encourages the uni-modal Gaussian posterior to move its mean close to the mean of the Gaussian prior, which makes it difficult for the Gaussian posterior to represent different modes conditioned on the observation.
%Similarly, this encourages the piecewise constant posterior to be similar to the piecewise constant prior.
%However, since the piecewise constant posterior is multi-modal, it may be able to shift some of its probability mass towards the prior distribution while keeping other probability mass on one or several modes dependent upon the output observation (e.g.\@ if the prior distribution is a uniform distribution and the true posterior concentrates all its probability mass in several small regions, then the approximate posterior could interpolate between the prior and the true posterior).

\section{Appendix: VHRED Implementation}
\label{appendix_b}
As described in the model section, the probability distribution of the generative model factorizes as:
\begin{align}
& P_\theta(\mathbf{w}_1, \ldots, \mathbf{w}_N) \nonumber \\ &= \prod_{n = 1}^N P_\theta(\mathbf{w}_n | \mathbf{w}_{< n}, z_n) P_{\theta}(z_n | \mathbf{w}_{< n}) \nonumber \\
& = \prod_{n = 1}^N \prod_{m = 1}^{M_n} P_\theta(w_{n, m} | w_{n, < m}, \mathbf{w}_{< n}, z_n) P_{\theta}(z_n | \mathbf{w}_{< n}),
\end{align}
where $\theta$ are the model parameters.
VHRED uses three RNN modules: an \textit{encoder} RNN, a \textit{context} RNN and a \textit{decoder} RNN.
First, each utterance is encoded into a vector by the \textit{encoder} RNN:
\begin{align*}
h^{\text{enc}}_{n, 0} = \mathbf{0}, \ \ h^{\text{enc}}_{n, m} = f^{\text{enc}}_{\theta}(h^{\text{enc}}_{n, m-1}, w_{n, m}) \\  \forall m=1,\dots,M_n,
\end{align*}
where $f^{\text{enc}}_{\theta}$ is either a GRU or a bidirectional GRU function.
The last hidden state of the \textit{encoder} RNN is given as input to the \textit{context} RNN.
The \textit{context} RNN uses this state to updates its internal hidden state:
\begin{align*}
h^{\text{con}}_{0} = \mathbf{0}, \ \ h^{\text{con}}_{n} = f^{\text{con}}_{\theta}(h^{\text{con}}_{n-1}, h^{\text{enc}}_{n, M_n}),
\end{align*}
where $f^{\text{con}}_{\theta}$ is a GRU function taking as input two vectors.
This state conditions the prior distribution over $z_n$:
\begin{align}
P_{\theta}(z_n \mid 
\mathbf{w}_{<n}) = f^{\text{prior}}_\theta (z_n ; h_{n-1}^{con}),
\end{align}
where $f^{\text{prior}}$ is a PDF parametrized by both $\theta$ and $h_{n-1}^{con}$.
Next, a sample is drawn from this distribution: $z_n \sim P_\theta (z_n | \mathbf{w}_{<n})$.
The sample and \textit{context} state are given as input to the \textit{decoder} RNN:
\begin{align*}
h^{\text{dec}}_{n, 0} = \mathbf{0}, \ \ & h^{\text{dec}}_{n, m} = f^{\text{dec}}_{\theta}(h^{\text{dec}}_{n, m-1}, h^{\text{con}}_{n-1}, z_{n}, w_{n, m}) \\ 
& \forall m=1,\dots,M_n,
\end{align*}
where $f^{\text{dec}}_{\theta}$ is the LSTM gating function taking as input four vectors.
The output distribution is computed by passing $h^{\text{dec}}_{n, m}$ through an MLP $f^{\text{mlp}}_{\theta}$, an affine transformation and a softmax function:
\begin{align}
& P_\theta(w_{n, m+1} | w_{n, \leq m}, \mathbf{w}_{< n}, z_n) \nonumber \\ & = \dfrac{e^{(O w_{n, m+1})^{\text{T}}f^{\text{mlp}}_{\theta}(h^{\text{dec}}_{n, m})}}{\sum_{w'}  e^{(O w')^{\text{T}}f^{\text{mlp}}_{\theta}(h^{\text{dec}}_{n, m})}}, \label{eq:hred_decoder_output}
\end{align}
where $O \in \mathbb{R}^{|V| \times d}$ is the word embedding matrix for the output distribution with embedding dimensionality $d \in \mathbb{N}$.

As mentioned in the model section, the approximate posterior is conditioned on the \textit{encoder} RNN state of the next utterance: 
\begin{align}
Q_{\psi}(z_n \mid \mathbf{w}_{\leq n}) = f^{\text{post}}_\psi (z_n ; h_{n-1}^{con}, h_{n, M_n}^{enc}),
\end{align}
where $f^{\text{post}}$ is a PDF parametrized by $\psi$ and $h_{n, M_n}^{enc}$ (i.e.\@ the future state of the \textit{encoder} RNN after processing $\mathbf{w}_n$).

For the Gaussian latent variables, we use the interpolation gating mechanism described in the main text for the approximate posterior. We experimented with other mechanisms for controlling the gating variables, such as defining  $\alpha_{\mu}$ and $\alpha_{\sigma}$ to be a linear function of the encoder. However, this did not improve performance in our preliminary experiments.

\section{Appendix: Training Details}
\label{appendix_d}

\textbf{Piecewise Constant Variable Interpolation}
We conducted initial experiments with the interpolation gating mechanism for the approximate posterior of the piecewise constant latent variables.
However, we found that this did not improve performance.
%As described in Sections \ref{nvdm} and \ref{vhred}, for all models that used piecewise latent variables, we chose to fix $\alpha_{a_i} = 1$, meaning the piecewise prior and posterior models are kept separate --- as opposed to having the posterior be an interpolation between another distribution and the prior --- since we found this to perform better. We believe that if $\alpha_{a_i} = 0$ for a long period of time, then the posterior receives no gradient signal. Without a gradient signal, the estimated posterior becomes increasingly disconnected from the rest of the model and, thus, less effective. This might be due to the choice of non-linearities, which affect the piecewise latent variables more so than the Gaussian latent variables.

\textbf{Dialogue Modeling} We use the Ubuntu Dialogue Corpus v2.0 extracted January, 2016: \url{http://cs.mcgill.ca/~jpineau/datasets/ubuntu-corpus-1.0/}.

For the \textit{HRED} model we found that an additional rectified linear units layer decreased performance on the validation set according to the activity F1 metric. Hence we test \textit{HRED} without the rectified linear units layer.
On the other hand, for all \textit{VHRED} models we found that the additional rectified linear units layer improved performance on the validation set.
For \textit{P-VHRED}, we found that a final weight of one for the KL divergence terms performed best on the validation set.
For \textit{G-VHRED} and \textit{H-VHRED}, reweighing the KL divergence terms with a final value $0.25$ performed best on the validation set.
We conducted preliminary experiments with $n=3$ and $n=5$ pieces, and found that models with $n=3$ were easier to train.
Therefore, we use $n=3$ pieces for both \textit{P-VHRED} and \textit{H-VHRED}.

For all models, we compute the log-likelihood and variational lower-bound costs starting from the second utterance in each dialogue.

\section{Appendix: Additional Document Modeling Experiments}
\label{appendix_e}

\textbf{Iterative Inference}
For the document modeling experiments, our results and conclusions depend on how tight the variational lower-bound is.
As such, it is in theory possible that some of our models are performing much better than reported by the variational lower-bound on the test set.
Therefore, we use a non-parametric iterative inference procedure to tighten the variational lower-bound, which aims to learn a separate approximate posterior for each test example.
The iterative inference procedure consists of simple stochastic gradient descent (no more than 100 steps), with a learning rate of $0.1$ and the same gradient rescaling used in training.
For 20 News-Groups, the iterative inference procedure is stopped on a test example if the bound does not improve over 10 iterations.
For Reuters and CADE, the iterative inference procedure is stopped if the bound does not improve over $5$ iterations.
During iterative inference the parameters of the model, as the well as the generated prior, are all fixed. 
Only the gradients of the variational lower-bound with respect to generated posterior model parameters (i.e.\@ the mean and variance of the Gaussian variables, and the piecewise components, $a_i$) are used to update the posterior model for each document (using a freshly drawn sample for each inference iteration step).

Note, this form of inference is expensive and requires additional meta-parameters (e.g.\@ a step-size and an early-stopping criterion).
We remark that a simpler, and more accurate, approach to inference might perhaps be to use importance sampling.

The results based on iterative inference are reported in Table \ref{ppl_results}. As Section \ref{doc_model}, we find that \textit{H-NVDM} outperforms the \textit{G-NVDM} model. This confirms our previous conclusions.

In our current examples, it appears that the \emph{H-NVDM} with 5 pieces returns more general words. For example, as evidenced in Table \ref{word_query_2}, in the case of ``government'', the baseline seems to value the plural form of the word (which is largely based on morphology) while the hybrid model actually pulls out meaningful terms such as ``federal'', ``policy'', and ``administration''. 

\begin{table}[!t]
%% increase table row spacing, adjust to taste
%\renewcommand{\arraystretch}{1.4}do
\caption{Word query similarity test on 20 News-Groups: for the query `government''.} %more general/abstract terms to the query, which may or may not always be what is desired.}
\label{word_query_2}
\centering
\small
\begin{tabular}{lllll}
%\hline\hline
\multicolumn{1}{l}{\begin{tabular}[x]{@{}c@{}}\textbf{G-NVDM}\\\end{tabular}}&\multicolumn{1}{c}{\begin{tabular}[x]{@{}c@{}}\textbf{H-NVDM-3}\\\end{tabular}}&\multicolumn{1}{c}{\begin{tabular}[x]{@{}c@{}}\textbf{H-NVDM-5}\\\end{tabular}}\tabularnewline
\hline
governments & citizens & arms \tabularnewline
citizens & rights & rights \tabularnewline
country & governments & federal \tabularnewline
threat & civil & country \tabularnewline
private & freedom & policy \tabularnewline
rights & legitimate & administration \tabularnewline
individuals & constitution & protect \tabularnewline
military & private & private \tabularnewline
freedom & court & citizens \tabularnewline
foreign & states & military \tabularnewline
\hline
\end{tabular}
\end{table}

\begin{table*}[!t]
%% increase table row spacing, adjust to taste
%\renewcommand{\arraystretch}{1.4}do
\caption{Comparative test perplexities on various document datasets (50 latent variables). Note that document probabilities were calculated using 10 samples to estimate the variational lower-bound.}
\label{ppl_results}
\centering
\begin{floatrow}
%\resizebox{8.5cm}{!} {%
\begin{tabular}{lrr}
%\hline\hline
\multicolumn{1}{l}{\begin{tabular}[x]{@{}c@{}}\textbf{20-NG}\\\end{tabular}}&\multicolumn{1}{c}{\begin{tabular}[x]{@{}c@{}}\textbf{Sampled}\\\end{tabular}}&\multicolumn{1}{c}{\begin{tabular}[x]{@{}c@{}}\textbf{SGD-Inf}\\\end{tabular}}\tabularnewline
\hline
%\textit{dAE} & $--$ & $--$ & $--$ & $--$\tabularnewline
\textit{LDA} & $1058$ & $--$ \tabularnewline
\textit{RSM} & $953$ & $--$ \tabularnewline
\textit{docNADE} & $896$ & $--$\tabularnewline
\textit{SBN} & $909$ & $--$ \tabularnewline
\textit{fDARN} & $917$ & $--$ \tabularnewline 
\textit{NVDM} & $836$ & $--$ \tabularnewline \hline
\textit{G-NVDM} & $651$ & $588$\tabularnewline 
\textit{H-NVDM-3} & $607$ & $546$ \tabularnewline
\textit{H-NVDM-5} & $\mathbf{566}$ & $\mathbf{496}$ \tabularnewline
%\textit{HC-NVDM-5} & $673$ & $596$\tabularnewline
\hline
\end{tabular}
%}
\begin{tabular}{lrr}
%\hline\hline
\multicolumn{1}{l}{\begin{tabular}[x]{@{}c@{}}\textbf{RCV1}\\\end{tabular}}&\multicolumn{1}{c}{\begin{tabular}[x]{@{}c@{}}\textbf{Sampled}\\\end{tabular}}&\multicolumn{1}{c}{\begin{tabular}[x]{@{}c@{}}\textbf{SGD-Inf}\\\end{tabular}}\tabularnewline
\hline
\textit{G-NVDM} & $905$ & $837$\tabularnewline 
\textit{H-NVDM-3} & $865$ & $807$ \tabularnewline 
%\textit{HC-NVDM-3} & $$ & $$\tabularnewline
\textit{H-NVDM-5} & $\mathbf{833}$ & $\mathbf{781}$ \tabularnewline
%\textit{HC-NVDM-5} & $$ & $$\tabularnewline
\textbf{} &  & \tabularnewline
\hline
\textbf{CADE} & \textbf{Sampled} & \textbf{SGD-Inf} \tabularnewline
\hline % SGD-inf patience of 5
\textit{G-NVDM} & $339$ & $230$\tabularnewline 
\textit{H-NVDM-3} & $\mathbf{258}$ & $\mathbf{193}$ \tabularnewline 
\textit{H-NVDM-5} & $294$ & $209$ \tabularnewline
\hline
\end{tabular}

\end{floatrow}
\end{table*}
% Table 1: Definitions of Sampled and SGD-Inf should appear in the caption. The table should stand alone.

\begin{table*}[t] %[!t]
%% increase table row spacing, adjust to taste
%\renewcommand{\arraystretch}{1.4}do
\caption{Approximate posterior word encodings (20-NG). For P-KL, we bold every case where piecewise variables showed greater word sensitivity than Gaussian variables w/in the same hybrid model.}
\label{doc_model_analysis}
\centering
\begin{floatrow}
\resizebox{7cm}{!} {%
\begin{tabular}{cccc}
%\hline\hline
\multicolumn{1}{c}{\begin{tabular}[x]{@{}c@{}}\textbf{Word}\\\end{tabular}}&\multicolumn{1}{c}{\begin{tabular}[x]{@{}c@{}}\textbf{G-NVDM}\\\end{tabular}}&\multicolumn{1}{c}{\begin{tabular}[x]{@{}c@{}}\textbf{H-NVDM-5}\\\end{tabular}}&\multicolumn{1}{c}{\begin{tabular}[x]{@{}c@{}}\textbf{}\\\end{tabular}}\tabularnewline
\textbf{Time-related} & \textbf{G-KL} & \textbf{G-KL} & \textbf{P-KL} \tabularnewline
\hline
months & 23 & 33 & \textbf{40} \tabularnewline
day & 28 & 32 & \textbf{35} \tabularnewline
time & \textbf{55} & 22 & \textbf{40} \tabularnewline
century & \textbf{28} & 13 & \textbf{19} \tabularnewline
past & \textbf{30} & 18 & \textbf{28} \tabularnewline
days & \textbf{37} & 14 & \textbf{19} \tabularnewline
ahead & \textbf{33} & 20 & \textbf{33} \tabularnewline
years & \textbf{44} & 16 & \textbf{38} \tabularnewline
today & 46 & 27 & \textbf{71} \tabularnewline
back & 31 & 30 & \textbf{47} \tabularnewline
future & \textbf{20} & 15 & \textbf{20} \tabularnewline
order & \textbf{42} & 14 & \textbf{26} \tabularnewline
minute & 15 & 34 & \textbf{40} \tabularnewline
began & \textbf{16} & 5 & \textbf{13} \tabularnewline
night & \textbf{49} & 12 & \textbf{18} \tabularnewline
hour & \textbf{18} & \textbf{17} & 16 \tabularnewline
early & 42 & 42 & \textbf{69} \tabularnewline
yesterday & 25 & 26 & \textbf{36} \tabularnewline
year & \textbf{60} & 17 & \textbf{21} \tabularnewline
week & 28 & 54 & \textbf{58} \tabularnewline
hours & 20 & 26 & \textbf{31} \tabularnewline
minutes & \textbf{40} & 34 & \textbf{38} \tabularnewline
months & 23 & 33 & \textbf{40} \tabularnewline
history & \textbf{32} & 18 & \textbf{28} \tabularnewline
late & 41 & \textbf{45} & 31 \tabularnewline
moment & \textbf{23} & \textbf{17} & 16 \tabularnewline
season & \textbf{45} & 29 & \textbf{37} \tabularnewline
summer & 29 & 28 & \textbf{31} \tabularnewline
start & 30 & 14 & \textbf{38} \tabularnewline
continue & 21 & 32 & \textbf{34} \tabularnewline
happened & 22 & 27 & \textbf{35} \tabularnewline
\hline
\end{tabular}
}

\resizebox{7cm}{!} {%
\begin{tabular}{cccc}
%\hline\hline
\multicolumn{1}{c}{\begin{tabular}[x]{@{}c@{}}\textbf{Word}\\\end{tabular}}&\multicolumn{1}{c}{\begin{tabular}[x]{@{}c@{}}\textbf{G-NVDM}\\\end{tabular}}&\multicolumn{1}{c}{\begin{tabular}[x]{@{}c@{}}\textbf{H-NVDM-5}\\\end{tabular}}&\multicolumn{1}{c}{\begin{tabular}[x]{@{}c@{}}\textbf{}\\\end{tabular}}\tabularnewline
\textbf{Names} & \textbf{G-KL} & \textbf{G-KL} & \textbf{P-KL} \tabularnewline
\hline
henry & 33 & \textbf{47} & 39 \tabularnewline
tim & \textbf{32} & 27 & 11 \tabularnewline
%patrick & VAL & 31 & 26 \tabularnewline
%dod & VAL & 171 & 62 \tabularnewline
mary & 26 & \textbf{51} & 30 \tabularnewline
james & 40 & \textbf{72} & 30 \tabularnewline
jesus & 28 & \textbf{87} & 39 \tabularnewline
george & 26 & \textbf{56} & 29 \tabularnewline
keith & 65 & \textbf{94} & 61 \tabularnewline
kent & 51 & \textbf{56} & 15 \tabularnewline
chris & 38 & \textbf{55} & 28 \tabularnewline
thomas & 19 & \textbf{35} & 19 \tabularnewline
hitler & 10 & \textbf{14} & 9 \tabularnewline
paul & 25 & \textbf{52} & 18 \tabularnewline
mike & 38 & \textbf{76} & 40 \tabularnewline
bush & \textbf{21} & 20 & 14 \tabularnewline \tabularnewline
\textbf{Adjectives} & \textbf{G-KL} & \textbf{G-KL} & \textbf{P-KL} \tabularnewline
\hline
american & \textbf{50} & 12 & \textbf{40} \tabularnewline
german & \textbf{25} & 21 & \textbf{22} \tabularnewline
european & 20 & 17 & \textbf{27} \tabularnewline
muslim & 19 & 7 & \textbf{23} \tabularnewline
french & 11 & \textbf{17} & \textbf{17} \tabularnewline
canadian & \textbf{18} & 10 & \textbf{16} \tabularnewline
japanese & 16 & 9 & \textbf{24} \tabularnewline
jewish & \textbf{56} & 37 & \textbf{54} \tabularnewline
%palestinian & 10 & 5 & 6 \tabularnewline
english & 19 & 16 & \textbf{26} \tabularnewline
%armenian & VAL & 3 & 3 \tabularnewline
islamic & 14 & 18 & \textbf{28} \tabularnewline
israeli & \textbf{24} & 14 & \textbf{18} \tabularnewline
british & \textbf{35} & 15 & \textbf{17} \tabularnewline
russian & 14 & 19 & \textbf{20} \tabularnewline
%lebanese & \textbf{15} & 7 & \textbf{10} \tabularnewline
\hline
\end{tabular}
}

\end{floatrow}
\end{table*}

\textbf{Approximate Posterior Analysis} We present an additional analysis of the approximate posterior on 20 News-Groups, in order to understand what the models are capturing. 
For a test example, we calculate the squared norm of the gradient of the KL terms w.r.t.\@ the word embedding inputted to the approximate posterior model.
The higher the squared norm of the gradients of a word is, the more influence it will have on the posterior approximation (encoder model).
For every test example, we count the top $5$ words with highest squared gradients separately for the multivariate Gaussian and piecewise constant latent variables.\footnote{Our approach is equivalent to counting the top $5$ words with the highest L2 gradient norms.}

%needed to formulate word scores, we follow the approach described in Sub-section \ref{dialog_model}, however, conditioning only on the (training) document bag-of-words to compute the latent posterior to then calculate the gradient of the KL-terms with respect to each word in the document.%\footnote{Since the vocabulary for 20-NG is relatively manageable, we opted for fully propagating the error gradients back to the input units.}

The results shown in Table \ref{doc_model_analysis}, illustrate how the piecewise variables capture different aspects of the document data.
The Gaussian variables were originally were sensitive to some of the words in the table.
However, in the hybrid model, nearly all of the temporal words that the Gaussian variables were once more sensitive to now more strongly affect the piecewise variables, which themselves also capture all of the words that were originally missed
This shift in responsibility indicates that the piecewise constant variables are better equipped to handle certain latent factors.
This effect appears to be particularly strong in the case of certain nationality-based adjectives (e.g., ``american'', ``israeli'', etc.).
While the \textit{G-NVDM} could model multi-modality in the data to some degree, this work would be primarily done in the model's decoder. In the \textit{H-NVDM}, the piecewise variables provide an explicit mechanism for capturing modes in the unknown target distribution, so it makes sense that the model would learn to use the piecewise variables instead, thus freeing up the Gaussian variables to capture other aspects of the data, as we found was the case with names (e.g., ``jesus'', ``kent'', etc.).

\section{Appendix: Additional Dialogue Modeling Experiments}
\label{appendix_f}

\begin{table*}[ht]
 \caption{Ubuntu model examples. The $\rightarrow$ token indicates a change of turn.}
 \label{table:ubuntu-examples}
 \scriptsize
 \centering
 \begin{tabular}{p{70mm}p{55mm}}
 \textbf{Dialogue Context (History)} & \textbf{Response} \\ \hline
Hi . I am installing ubuntu now in my new laptop . In " something else " partitioning , what mount point should I set for a drive which is not root or not home ...  $\rightarrow$ It 's up to you , just choose a directory that will remind you of the contents of that partition . E.G. : if it 's the Windows partition , use /windows .  $\rightarrow$ it 's a new harddrive with full free space . I bought it without windows preinstalled .  I want to create drives in which I will only store files .. I mean , not root or not home . What mount point do I set for it ? " /mount " is not shown in drop down menu  sorry . I mean /mount  I mean , in my desktop , extra drives are mounted in /media  do you understand my problem ? Sorry , English is not my native language .  $\rightarrow$ I do :) Just use a directory with a meaningful name and preferably in the root , for example " /files ".  Choose something like " /files ".  $\rightarrow$ / for root , /home for home , but what mount point for a file storage partition .  wait , let me see if there is a " /files " , thank you .  & \textbf{HRED:} No problem . \newline \textbf{G-VHRED:} It should be in the list of the drive . If you want to be able to mount it .  \newline \textbf{P-VHRED:} If you want to mount it , you can use the mount command .  \newline \textbf{H-VHRED:} As far as I know , you can use the " mount " command to mount the partitions .  \\ \hline
I have a problem installing Ubuntu 12.04 desktop . Tried desktop with USB -> hangs . alternate with usb -> hangs . alternate with cd -> hangs ( even faster) . I tried using different usb drives . Could someonle please help me ?  $\rightarrow$ how are you making the usb ? and whats the pc specs .  $\rightarrow$ I used Universal usb installer  $\rightarrow$ what exactly does it do when it ' hangs '  $\rightarrow$ Specs : CPU core i7 @ 3.4 Ghz , Sapphire Radeon HD 7800 , 8GB RAM  $\rightarrow$ ATi video ? tried the NOMODESET option yet ?  $\rightarrow$ nothing at all , I also can't witch to console using Ctrl + alt + F  not yet  Will try that , thanks  & \textbf{HRED:} you could try the alternate install cd .  \newline \textbf{G-VHRED:} it might be a problem with your graphics card . You could also try the alternate install cd and see if it works . \newline \textbf{P-VHRED:} try ' nomodeset ' or ' nomodeset ' \newline \textbf{H-VHRED:} only thing you can do is run the command ' sudo dpkg-reconfigure xserver-xorg ' and see if you get any errors .  \\ \hline
anyone know how to add a route in ubuntu ?  getting an error on adding a route , and I'm not sure why  $\rightarrow$ pastebin your command and pronlem  $\rightarrow$ \textless url \textgreater  & \textbf{HRED:} can you pastebin the output of " route -n " please \newline \textbf{G-VHRED:} if you want to use pastebin to pastebin the output of " sudo route -n " please pastebin the output of " route " \newline \textbf{P-VHRED:} what is the output of : route -n \newline \textbf{H-VHRED:} only if you have a static ip address , you need to specify the ip address of the gateway  \\ \hline
Any idea why empathy 's not playing notification sounds ? Even though I have ' em ticked in preferences  $\rightarrow$ restarted it yet ?  $\rightarrow$ yar  $\rightarrow$ check pulse to see if the application is muted for some reason ?  well Sound settings .  $\rightarrow$ Had sound effects turned off in sound settings , didn't realize that controlled other applications  $\rightarrow$ Ah yea , ive done it a few time it 's annoying  My favorite though is recently pulse has been freezing on my desktop and audio will just not be adjustable for like ... 30 seconds or so  & \textbf{HRED:} I don't think it 's a problem with pulseaudio , but I'm not sure how to fix it \newline \textbf{G-VHRED:} Thank you for your help . I will try that . Thanks for the suggestion . \newline \textbf{P-VHRED:} I think it 's a bug with pulseaudio , but I don't know how to fix it . \newline \textbf{H-VHRED:} as far as I can tell , I have a problem with pulseaudio , but I don't know how to fix it .  \\ \hline
Question about setting up drives : Running 12.04 LTS . I have two drives . An SSD that holds the OS for booting , and a second drive " data " for all the data , home directories , etc . The data drive is encrypted with LUKS . With my current setup , whenever I reboot , I get a password entry screen to decrypt the data drive , where my home directory is . This works fine . The problem is that if I reboot remotely , I'm stuck as I can't enter  password . If I keep my home directory on the SSD , there won't be enough room . If I use a symbolic link , and the encrypted drive isn't mounted , that creates errors and issues . Suggestions ??  $\rightarrow$ yes , you can redirect the  $\rightarrow$ Message was cut off ...  & \textbf{HRED:} if you want to see the data on the drive , you can mount it in /etc/fstab \newline \textbf{G-VHRED:} it should be a good place to do that , but if you want to do a clean install , you will need to reboot\newline \textbf{P-VHRED:} it 's not a problem , it 's just a bug \newline \textbf{H-VHRED:} if you want to see the data on the drive , you can mount it in /etc/fstab  \\ \hline
hey guys , does anyone know why i cannot connect to ssh via my external IP ? (connection refused ) but ssh -v localhost works fine ?  $\rightarrow$ Have you forwarded the port on your router to the machine you're trying to access ?  $\rightarrow$ It is a work machine , so am not sure of the rules & \textbf{HRED:} You need to forward port 22 to your router to forward port 22 to the server . \newline \textbf{G-VHRED:} That 's odd . What are you trying to do ?  Can you pastebin the output of " sudo netstat " to " pastebin " please ? \newline \textbf{P-VHRED:} Can you pastebin the output of " sudo apt-get install openssh-server "? \newline \textbf{H-VHRED:} Even if it 's not working , then you need to set the port forward to your router . \\ \hline
 \end{tabular}
\end{table*}

{\bf Ubuntu Experiments } We present test examples --- dialogue context and model responses generated using beam search --- for the Ubuntu models in Table \ref{table:ubuntu-examples}. The examples qualitatively illustrate the differences between models.
First, we observe that \textit{HRED} tends to generate highly generic responses compared to all the latent variable models. This supports the quantitative results reported in the main text, and suggests that modeling the latent factors through latent variables is critical for this task.
Next, we observe that \textit{H-VHRED} tends to generate relevant entities and commands --- such as \textit{mount command}, \textit{xserver-xorg}, \textit{static ip address} and \textit{pulseaudio} in examples 1-4.
On the other hand, \textit{G-VHRED} tends to be better at generating appropriate verbs --- such as \textit{list}, \textit{install}, \textit{pastebin} and \textit{reboot} in examples   1-3 and example 5.
Qualitatively, \textit{P-VHRED} model appears to perform somewhat worse than both \textit{G-VHRED} and \textit{H-VHRED}. This suggests that the Gaussian latent variables are important for the Ubuntu task, and therefore that the best performance may be obtained by combining both Gaussian and piecewise latent variables together in the \textit{H-VHRED} model.

%{\bf Approximate Posterior Analysis    }
{\bf Twitter Experiments } We also conducted a dialogue modeling experiment on a Twitter corpus, extracted from based on public Twitter conversations \citep{ritter2011data}.
The dataset is split into training, validation, and test sets, containing respectively 749,060, 93,633 and 9,399 dialogues each.
On average, each dialogue contains about $6$ utterances (dialogue turns) and about $94$ words.
We pre-processed the tweets using byte-pair encoding \citep{sennrich2015neural} with a vocabulary consisting of 5000 sub-words.

We trained our models with a learning rate of $0.0002$ and mini-batches of size $40$ or $80$.\footnote{We had to vary the mini-batch size to make the training fit on GPU architectures with low memory.}
As for the Ubuntu experiments, we used a variant of truncated back-propagation and apply gradient clipping.
We experiment with \textit{G-VHRED} and \textit{H-VHRED}. 
Similar to \citep{serban2016hierarchical}, we use a bidirectional GRU RNN \textit{encoder}, where the forward and backward RNNs each have $1000$ hidden units.
We experiment with \textit{context} RNN encoders with $500$ and $1000$ hidden units, and find that that $1000$ hidden units reach better performance w.r.t.\@ the variational lower-bound on the validation set.
The \textit{encoder} and \textit{context} RNNs use layer normalization \citep{ba2016layer}.
We experiment with \textit{decoder} RNNs with $1000$, $2000$ and $4000$ hidden units (LSTM cells), and find that $2000$ hidden units reach better performance.
For the \textit{G-VHRED} model, we experiment with latent multivariate Gaussian variables with $100$ and $300$ dimensions, and find that $100$ dimensions reach better performance.
For the \textit{H-VHRED} model, we experiment with latent multivariate Gaussian and piecewise constant variables each with $100$ and $300$ dimensions, and find that $100$ dimensions reach better performance.
We drop words in the decoder with a fixed drop rate of $25\%$ and multiply the KL terms in the variational lower-bound by a scalar, which starts at zero and linearly increases to $1$ over the first 60,000 training batches.
Note, unlike the Ubuntu experiments, the final weight of the KL divergence is exactly one (hence the bound is tight).

%We also experiment with an \textit{HRED} model. For this model, we use the same \textit{encoder} and \textit{context} RNN architectures as the \textit{G-VHRED} and \textit{H-VHRED} models described above. We set the \textit{encoder} RNN to have $1000$ hidden units.

\newcommand{\heart}{\ensuremath\heartsuit}

\begin{table*}[t]
  \caption{Approximate posterior word encoding on Twitter. The numbers are computed by counting the number of times each word is among the $5$ words with the largest sum of squared gradients of the Gaussian KL divergence (G-KL) and piecewise constant KL divergence (P-KL)}
  \label{tabel:approximate_posterior_word_sensitivity}
  \small
  \centering
    \begin{tabular}{c c c c c c c c }
         
    \textbf{Word} & \textbf{G-VHRED} & \multicolumn{2}{c}{\textbf{H-VHRED}} & \textbf{Word} & \textbf{G-VHRED} & \multicolumn{2}{c}{\textbf{H-VHRED}} \\ 
    \textbf{Time-related} & \textbf{G-KL} & \textbf{G-KL} & \textbf{P-KL} & \textbf{Event-related} & \textbf{G-KL} & \textbf{G-KL} & \textbf{P-KL} \\
    \hline
    monday & 3 & 5 & \textbf{10} & school & 9 & 16 & \textbf{50} \\
    tuesday & 2 & 3 & \textbf{7} & class & 11 & 16 & \textbf{27} \\
    wednesday & 4 & 11 & \textbf{13} & game & 20 & 26 & \textbf{41} \\
    thursday & 2 & 3 & \textbf{9} & movie & 12 & 20 & \textbf{41} \\
    friday & 9 & 18 & \textbf{26} & club & 13 & 22 & \textbf{28} \\
    saturday & 6 & 6 & \textbf{13} & party & 8 & 10 & \textbf{32} \\
    sunday & 2 & 2 & \textbf{9} & wedding & 7 & 13 & \textbf{23} \\
    weekend & 8 & 16 & \textbf{32} & birthday & 12 & 20 & \textbf{23} \\
    today & 18 & 28 & \textbf{56} & easter & 15 & 15 & \textbf{23} \\
    night & 16 & 31 & \textbf{68} & concert & 7 & 16 & \textbf{20} \\
    tonight & 32 & 36 & \textbf{47} & dance & 11 & 12 & \textbf{21} \\
    & & & & & & & \\
    \textbf{Word} & \textbf{G-VHRED} & \multicolumn{2}{c}{\textbf{H-VHRED}} & \textbf{Word} & \textbf{G-VHRED} & \multicolumn{2}{c}{\textbf{H-VHRED}} \\ 
    \parbox[c][2.00em][c]{0.1\textwidth}{\textbf{Sentiment \\ -related}} & \textbf{G-KL} & \textbf{G-KL} & \textbf{P-KL} & \parbox[c][2.00em][c]{0.235\textwidth}{\textbf{Acronyms, Punctuation \\ Marks \& Emoticons}} & \textbf{G-KL} & \textbf{G-KL} & \textbf{P-KL} \\ \hline
  good & \textbf{72} & \textbf{73} & 44 & lol & \textbf{394} & \textbf{358} & 312 \\
  love & \textbf{102} & \textbf{101} & 38 & omg & \textbf{52} & \textbf{45} & 19 \\
  awesome & \textbf{26} & \textbf{44} & 39 & . & 386 & 558 & \textbf{1009} \\
  cool & 14 & 28 & \textbf{29} & ! & \textbf{648} & \textbf{951} & 525 \\
  haha & \textbf{132} & \textbf{101} & 75 & ? & \textbf{507} & \textbf{851} & 221 \\
  hahaha & \textbf{60} & \textbf{48} & 24 & * & \textbf{108} & \textbf{54} & 19 \\
  amazing & \textbf{14} & \textbf{38} & 33 & xd & \textbf{28} & \textbf{42} & 26 \\
  thank & \textbf{137} & \textbf{153} & 29 & \heart & \textbf{56} & \textbf{42} & 24 \\ \hline
    \end{tabular}
\end{table*}

Our hypothesis is that the piecewise constant latent variables are able to capture multi-modal aspects of the dialogue.
Therefore, we evaluate the models by analyzing what information they have learned to represent in the latent variables.
For each test dialogue with $n$ utterances, we condition each model on the first $n-1$ utterances and compute the latent posterior distributions using all $n$ utterances.
We then compute the gradients of the KL terms of the multivariate Gaussian and piecewise constant latent variables w.r.t.\@ each word in the dialogue.
Since the words vectors are discrete, we compute the sum of the squared gradients w.r.t.\@ each word embedding.
% Alex: should we mention we normalize to ensure values sum to 1?
% Julian: it doesn't matter for this result. Since we normalize across words, the top 5 most sensitive words are still the same after normalization as before normalization. But I will add a note in the 
The higher the sum of the squared gradients of a word is, the more influence it will have on the posterior approximation (encoder model).
For every test dialogue, we count the top $5$ words with highest squared gradients separately for the multivariate Gaussian and piecewise constant latent variables.\footnote{Our approach is equivalent to counting the top $5$ words with the highest L2 gradient norms. We also did some experiments using L1 gradient norms, which showed similar patterns.}

The results are shown in Table \ref{tabel:approximate_posterior_word_sensitivity}.
The piecewise constant latent variables clearly capture different aspects of the dialogue compared to the Gaussian latent variables.
The piecewise constant variable approximate posterior encodes words related to time (e.g.\@ weekdays and times of day) and events (e.g.\@ parties, concerts, Easter).
On the other hand, the Gaussian variable approximate posterior encodes words related to sentiment (e.g.\@ laughter and appreciation) and acronyms, punctuation marks and emoticons (i.e.\@ smilies). We also conduct a similar analysis on the document models evaluated in Sub-section \ref{doc_model}, the results of which may be found in the Appendix.

\end{document}